\documentclass[oneside, 11pt]{article}

\usepackage[T1]{fontenc}
\usepackage[utf8]{inputenc}
\usepackage[sc]{mathpazo}  %
\usepackage{jmlr2e_cus}
\usepackage[usenames]{xcolor}
\usepackage[bookmarksnumbered=true]{hyperref}
\usepackage{url}
\definecolor{cite_color}{HTML}{114083}
\definecolor{link_color}{RGB}{153, 0,0}  %
\definecolor{url_color}{RGB}{153, 102,  0}
\definecolor{emp_color}{RGB}{0,0,255}
\hypersetup{
 colorlinks,
 citecolor=cite_color,
 linkcolor=link_color,
 urlcolor=url_color}
\graphicspath{{./figs/}}

\usepackage{framed}
\definecolor{shadecolor}{rgb}{0.94, 0.97, 1.0}

\usepackage{epsfig}
\usepackage{amsmath}
\usepackage{amssymb}
\usepackage{mathrsfs}
\usepackage{graphicx}
\usepackage{subfig}
\usepackage{multirow}
\usepackage{booktabs}
\usepackage{wrapfig}
\usepackage{tikz,siunitx,mwe}
\usepackage{caption}
\usepackage{enumerate}
\usepackage{bm}
\usepackage{tabularx}

\usepackage{natbib}
\bibpunct{(}{)}{;}{a}{,}{,}
\usepackage{mathtools}

\usepackage[vlined, ruled, linesnumbered]{algorithm2e}

\SetCommentSty{mycommfont}
\SetKwInOut{Input}{input}
\SetKwInOut{Output}{output}

 \usepackage{cleveref} %
 \crefname{section}{Section}{Sections}
 \crefname{theorem}{Theorem}{Theorems}
 \crefname{lemma}{Lemma}{Lemmas}
 \crefname{equation}{Equation}{Equations}
 \crefname{proposition}{Proposition}{Propositions}
 \crefname{claim}{Claim}{Claims}
\crefname{appendix}{Appendix}{Appendices}
   \crefname{algorithm}{Algorithm}{Algorithms}
 \crefname{figure}{Figure}{Figures}
 \crefname{table}{Table}{Tables}
 \crefname{remark}{Remark}{Remarks}
 \crefname{definition}{Definition}{Definitions}
 \crefname{equatinon}{Equation}{Equations}
 \crefname{corollary}{Corollary}{Corollaries}

\allowdisplaybreaks

\usepackage{thmtools}
\usepackage{thm-restate}
\let \oldtextcircled \textcircled
\renewcommand{\textcircled}[1]{\oldtextcircled{\footnotesize #1}}

\newcommand{\scalebarimg}[3]{
  \begin{tikzpicture}
    \draw node[name=micrograph] {\includegraphics[width=#2\textwidth]{#1}}; %
    \draw  (micrograph.north west)  node[anchor=north west,yshift=-1, ]{\textbf{\small{#3}}}; %
  \end{tikzpicture}
}

\usepackage{enumitem}
\setlist[itemize]{leftmargin=9mm}

\newcommand{\todo}[1]{\textcolor{blue}{TODO:#1}}
\newcommand{\appendixtitle}[1]{
	\begin{center}
		\LARGE \bf #1
	\end{center}
}

\def \x{\mathbf{x}}

 \firstpageno{1}
\usepackage{listings}
\usepackage{xcolor}
\usepackage{makecell}
\usepackage{amssymb}
\usepackage{amsmath}
\usepackage{amsfonts}
\usepackage{graphicx}
\usepackage{hyperref}
\usepackage{url}
\usepackage{verbatim}

\usepackage{subfloat}
\usepackage{pythonhighlight}

\definecolor{codegreen}{rgb}{0,0.6,0}
\definecolor{codegray}{rgb}{0.5,0.5,0.5}
\definecolor{codepurple}{rgb}{0.58,0,0.82}
\definecolor{backcolour}{rgb}{0.95,0.95,0.92}

\definecolor{lightgreen}{HTML}{A9D18E}
\definecolor{lightblue}{HTML}{9DC2E6}

\lstdefinestyle{mystyle}{
    backgroundcolor=\color{backcolour},   
    commentstyle=\color{codegreen},
    keywordstyle=\color{magenta},
    numberstyle=\tiny\color{codegray},
    stringstyle=\color{codepurple},
    basicstyle=\ttfamily\footnotesize,
    breakatwhitespace=false,         
    breaklines=true,                 
    captionpos=b,                    
    keepspaces=true,                 
    numbers=left,                    
    numbersep=5pt,                  
    showspaces=false,                
    showstringspaces=false,
    showtabs=false,                  
    tabsize=2
}

\usepackage[edges]{forest}
\usetikzlibrary{arrows.meta}
\usepackage{setspace}
\newcommand{\chembl}{ChEMBL\xspace}
\newcommand{\drugood}{\texttt{DrugOOD}\xspace}
\newcommand{\aidd}{{AIDD}\xspace}

\begin{document}

\title{{\LARGE
DrugOOD: Out-of-Distribution (OOD) Dataset Curator and Benchmark for AI-aided Drug Discovery}\\
 {\Large -- A Focus on Affinity Prediction Problems with Noise Annotations}
}

\author{
\normalfont{Yuanfeng Ji$^{1,3,*}$} \\
\And
\normalfont{Lu Zhang$^{1,2,}$\thanks{Equal contribution. Order was determined by tossing a coin.}} \\
\And
\normalfont{Jiaxiang Wu$^{1}$} \\
\And
\normalfont{Bingzhe Wu$^{1}$} \\
\And
\normalfont{Long-Kai Huang$^{1}$} \\
\And
\normalfont{Tingyang Xu$^{1}$} \\
\And
\normalfont{Yu Rong$^{1}$} \\
\And
\normalfont{Lanqing Li$^{1}$} \\
\And
\normalfont{Jie Ren$^{1}$} \\
\And
\normalfont{Ding Xue$^{1}$} \\
\And
\normalfont{Houtim Lai$^{1}$} \\
\And
\normalfont{Shaoyong Xu$^{1}$} \\
\And
\normalfont{Jing Feng$^{1}$} \\
\And
\normalfont{Wei Liu$^{1}$} \\
\And
\normalfont{Ping Luo$^{3}$} \\
\And
\normalfont{Shuigeng Zhou$^{2}$} \\
\And
\normalfont{Junzhou Huang$^{1}$} \\
\And
\normalfont{Peilin Zhao$^{1}$} \\
\And
\normalfont{Yatao Bian$^{1}$\thanks{Correspondence to: Yatao Bian, email: \texttt{DrugAIOOD@gmail.com}}} \\
\AND
{\normalfont $^1$Tencent AI Lab, China}\\
$^2$Fudan University, China \\
$^3$The University of Hong Kong, China\\[.5em]%
\today
\setcounter{footnote}{0}
}

\maketitle

\begin{abstract}%

AI-aided drug discovery (\aidd) is gaining increasing popularity due to its promise of making the search for new pharmaceuticals quicker, cheaper and more efficient.
In spite of its extensive use in many fields, such as ADMET prediction, virtual screening,  protein folding  and generative chemistry, little has been explored in terms of the out-of-distribution (OOD) learning problem with \emph{noise}, which is inevitable in real world \aidd applications. 

In this work, 
we present \drugood\footnote{Project Page: \url{https://drugood.github.io}}, a systematic OOD dataset curator and benchmark for AI-aided drug discovery,  which comes with an open-source Python package that fully automates the data curation and OOD benchmarking processes.   
We focus on one of the most crucial problems in \aidd: drug target binding affinity prediction, which involves both macromolecule (protein target) and small-molecule (drug compound). 
In contrast to only providing fixed datasets,  \drugood offers automated dataset curator with user-friendly customization scripts,  rich domain annotations aligned with biochemistry knowledge, realistic noise annotations and rigorous benchmarking of state-of-the-art OOD algorithms.
Since the molecular data is often modeled as irregular graphs using graph neural network (GNN) backbones, \drugood also serves as a valuable testbed for \emph{graph OOD learning} problems.  
Extensive empirical studies have shown a significant performance gap between in-distribution and out-of-distribution experiments, which  highlights the need to develop better schemes that can allow for OOD generalization under noise for \aidd.  
\end{abstract}

\vspace{.2cm}

\newpage 
{\footnotesize
\hypersetup{linkcolor=cite_color}
\tableofcontents
}
\clearpage

\section{Introduction}
\label{sec_intro}

The traditional drug discovery process is extremely time-consuming and expensive.  
Typically, the development of a new drug takes nearly a decade and costs about \$3 billion \citep{pushpakom2019drug}, whereas about 90\% of experimental drugs fail during lab, animal or human testing. 
Meanwhile, the number of drugs approved every year per dollar spent on development has plateaued or decreased for most of the past decade \citep{nosengo2016can}. 
To accelerate the development for new drugs, drugmakers and investors turn their attention to artificial intelligence \citep{muratov2020qsar} techniques for drug discovery, which aims at rapidly identifying new compounds and modeling complex mechanisms in the body to automate previously manual processes \citep{schneider2018automating}.

The applications of AI aided drug discovery is being continuously extended 
in the pharmaceutical field, ranging from ADMET prediction \citep{wu2018moleculenet,rong2020self}, target identification \citep{zeng2020target, mamoshina2018machine}, protein structure prediction and protein design \citep{jumper2021highly,baek2021accurate,gao2020deep},  retrosynthetic analysis \citep{coley2017computer,segler2018planning,yan2020retroxpert}, search of antibiotics \citep{stokes2020deep},  
generative chemistry \citep{sanchez2017optimizing, simonovsky2018graphvae}, drug repurposing for emerging diseases \citep{gysi2021network} to virtual screening \citep{hu2016large, karimi2019deepaffinity, lim2019predicting}.  
Among them, virtual screening is one of the most important yet challenging applications.
The aim of virtual screening is to pinpoint a small set of compounds with high binding affinity for a given target protein in the presence of a large number of candidate compounds.    
A crucial task in solving the virtual screening problem is to develop computational approaches to predict the binding affinity of a given drug-target pair, which is the main task studied in this paper.

In the field of AI-aided drug discovery, the problem of \emph{distribution shift},  where the training distribution differs from the test distribution, is ubiquitous.
For instance, when performing virtual screening for hit finding, the prediction model is typically trained on known target proteins.  
However, a ``black swan'' event like COVID-19 can occur, resulting in a new target with unseen data distribution.
The performance on the new target will significantly degrades.
To handle the performance degradation \citep{WILDS} caused by distribution shift, it is essential to develop robust and generalizable algorithms for this challenging setting in \aidd.   
Despite its importance in the real-world problem, curated OOD datasets and benchmarks are currently lacking in addressing generalization in AI-aided drug discovery. 

\begin{figure}[!t]
	\centering 
	\includegraphics[width=1\textwidth]{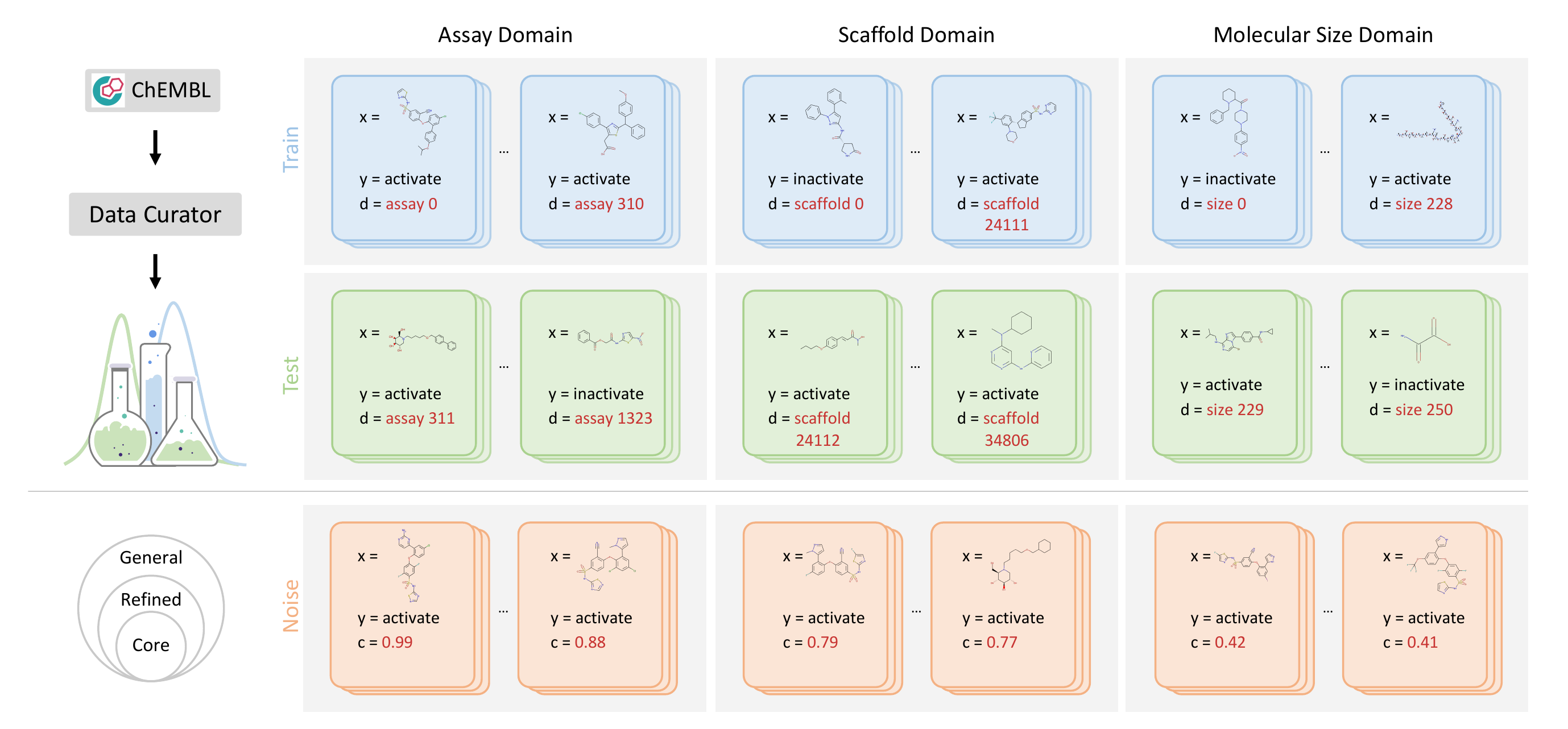}
	\caption{\drugood provides large-scale, realistic, and diverse datasets for Drug AI OOD research. Specifically, \drugood focuses on the problem of domain generalization, in which we train and test the model on disjoint domains, e.g., molecules in a new assay environment.
	\textit{Top Left:} Based on the ChEMBL database, we present an automated dataset curator for customizing OOD datasets flexibly.
	\textit{Top Right:} \drugood releases realized exemplar datasets spanning different domain shifts. In each dataset, each data sample $(\x, y, d)$ is associated with a domain annotation $d$. We use the background colours \textcolor{lightblue}{blue} and \textcolor{lightgreen}{green} to denote the seen data and unseen test data.
	\textit{Bottom}: Examples with different noise levels from the \drugood dataset. \drugood identifies and annotates three noise levels (left to right: core, refined, general) according to several criteria, and as the level increases,  data volume increases and  more noisy sources are involved.}
	\label{fig:overview}
\end{figure}

Another essential issue in the field of AI-aided drug discovery is the label noise.
The AI model are typical trained on public datasets, such as \chembl, whereas the bioassay data in the dataset are often noisy \citep{kramer2012experimental,cortes2016consistent}.
For example, the activity data provided in \chembl is extracted manually from full text articles in seven Medicinal Chemistry journals \citep{mendez2019chembl}. 
Various factors can cause noise in the data provided in \chembl, including but not limited to different confidence levels for activities measured through experiments, 
unit-transcription errors, repeated citations of single measurements and different ``cut-off'' noise\footnote{E.g., measurements could be recorded  with $<$, $\leq$,  $\approx$, >, $\geq$ etc, which would introduce the ``cut-off'' noise when translated into supervision labels.}.
\cref{fig:overview} shows examples with different noisy levels.
Meanwhile, real world data with noise annotations is lacking for learning tasks under noise labels \citep{angluin1988learning, han2020survey}. %

To help accelerate research by focusing community attention and simplifying systematic comparisons between data collection and implementation method, 
we present  \drugood,  a systematic OOD dataset curator and benchmark for AI-aided drug discovery which comes with an open-source Python package that fully automates the data curation process and OOD benchmarking process.   
We focus on the most challenging OOD setting: domain generalization \citep{zhou2021domain} problem in AI-aided drug discovery, though \drugood can be easily adapted to other OOD settings, such as subpopulation shift \citep{WILDS} and domain adaptation \citep{zhuang2020comprehensive}. Our dataset is also the first \aidd dataset curator with realistic noise annotations, that can serve as an important testbed for the setting of  learning  under noise.  

Notably, we present an automated dataset curator based on the large-scale bioassay deposition website \chembl \citep{mendez2019chembl}, in contrast to just providing a set of curated datasets. \cref{fig:intro:curator} gives an overview of the automated dataset curator.  
Using this dataset curator, potential researchers/practitioners can generate new OOD datasets based on their specific needs by simply re-configuring the curation process, i.e., modifying the YAML files in the python package. 
Specifically, we also realize this dataset curator by generating 96 OOD datasets spanning various domains, noise annotations and measurement types. This mechanism comes with two advantages:  i) It ensures that our released datasets and benchmarks are fully reproducible. ii) It allows great flexibility for future usage since it is often difficult, even for domain experts, to agree on one specific configuration.  
As an example, using EC50 as a measure of affinity, agreeing on a threshold for partitioning to find active/inactive pairs may be challenging.

As OOD learning subsumes or is closed related to  other learning settings with distribution shift, such as domain adaption \citep{zhuang2020comprehensive}, transfer learning \citep{pan2009survey}, and zero-shot learning \citep{romera2015embarrassingly, wang2019survey}, \drugood can also serve as a benchmark dataset or be used to generate datasets to study affinity prediction problems in \aidd under these learning settings.    
The following components summarize our major contributions:
\begin{enumerate}
\item \emph{Automated Dataset Curator}: We provide a fully customizable pipeline for curating OOD datasets  for AI-aided drug discovery from the large-scale bioassay deposition website \chembl. 

\item \emph{Rich domain annotations}:  We present various approaches to generate specific domains that are aligned with the domain knowledge of biochemistry. 

\item \emph{Realistic noise annotations}: We annotate real-world noise according to the measurement confidence score, ``cut-off'' noise etc, offering a valuable testbed for learning under real-world noise.

\item \emph{Rigorous OOD benchmarking}:  We benchmark six  SOTA OOD algorithms with various backbones for the 96 realized dataset instances and gain insight into OOD learning under noise for \aidd.      
\end{enumerate}

\paragraph{Paper Organizations.}  
\cref{sec_background_and_related_work} presents  background and related work on AI-aided drug discovery, existing OOD algorithms, datasets, benchmarks and  affinity prediction-related materials. In \cref{sec_curator} we provide details on the automated dataset curator with real-world domain and noise annotations.  We present specifics on benchmarking SOTA OOD algorithms
in \cref{sec_benchmark_sota_ood_alg}. \cref{sec_package_usage} gives implementation details and package usage guidelines.  
We present experimental results and corresponding discussions in \cref{sec_exps}.  
Lastly, \cref{sec_disc} discusses and concludes the paper.

\section{Background and Related Work}
\label{sec_background_and_related_work}

In this section, we review the current progress in binding affinity prediction problems, one of highly active research areas in AIDD. The performance of affinity prediction is often limited by OOD issues and noisy labels, which motivates us to propose the \drugood{} database to explicitly tackle such problems. Lastly, we summarize general methods for OOD and noisy labels, together with representation learning for affinity prediction for virtual screening, which are later used for benchmark tests on the \drugood{} datasets.

\subsection{Binding Affinity Prediction in AI-aided Drug Discovery}

The ability of AI techniques has been dramatically boosted in various domains, mainly due to wide-spread applications of deep neural networks. We have witnessed a growing number of researches attempting to solve traditional problems in the drug discovery with more advanced AI models. There have been several surveys \citep{Sliwoski2013, jing2018deep, yang2019concepts, paul2021artificial, deng2021artificial,bender2021artificial} summarizing recent advances and problems in this area, covering key aspects including major applications, representative techniques, and critical assessment benchmarks.

As pointed out in \citep{yang2019concepts, paul2021artificial}, most of AI-driven applications can be roughly categorized into two domains, {i.e.,} molecule generation and molecule screening. Molecule generation aims at adopting generative models to produce a large pool of candidate drug molecules with certain restraints satisfied \citep{simonovsky2018graphvae, sanchez2017optimizing, satorras2021en}. On the other hand, molecule screening attempts to identify the most promising molecule(s) based on a wide range of predicted properties \citep{yang2019analyzing, feinberg2020improvement, jimenez2018k}. Other typical applications of AI techniques in drug discovery include target identification \citep{zeng2020target, mamoshina2018machine}, target structure prediction \citep{jumper2021highly, baek2021accurate}, drug re-purposing \citep{aliper2016deep, issa2021machine, pham2021deep}, and molecule retrosynthesis \citep{coley2017computer, zheng2019predicting, chen2020retro}.

For conducting virtual screening on candidate molecules, both target-independent (\textit{e.g.} ADMET) and target-dependent (\textit{e.g.} binding affinity) properties are critical. The former ones measure how likely the molecule itself is qualified as a candidate drug, for instance, it should not induce severe liver toxicity to human \citep{zhang2016silico, asilar2020image}. On the other hand, target-dependent properties consider the tendency of its potential interaction with the target (and other unrelated proteins), which often heavily depends on the joint formulation of candidate molecule and target \citep{hu2016large, karimi2019deepaffinity, lim2019predicting}. In this paper, we mainly concentrate on the binding affinity between molecule and protein target, which falls into the domain of predicting target-dependent properties.
In this circumstance, the  out-of-distribution issue may result in severe performance degradation (e.g., when the target distribution dramatically differs between model training and inference), which leads to the major motivation of this paper.

\subsubsection{Databases}

ChEMBL \citep{davies2015chembl, mendez2019chembl} is a large-scale open-access database consists of small molecules and their biological activity data. Such information is mainly extracted from medicinal chemistry journal articles, supplemented with data collected from approved drugs and clinical development candidates. It now contains over 2.1 million distinct compounds and 18.6 million records of their activities, which involve over 14,500 targets.

BindingDB \citep{gilson2016bindingdb} collects experimental interaction data between proteins and small molecules, primarily from scientific articles and US patents. BindingDB also gathers selected data entries from other related databases, including PubChem \citep{wang2009pubchem}, \chembl \citep{mendez2019chembl}, PDSP Ki \citep{roth2000multiplicity}, and CSAR \citep{carlson2011call}. Advanced search tools, hypothesis generation schemes (from targets to compounds and vice versa), and virtual compound screening methods are also integrated in the database.

PDBbind \citep{liu2014pdb} is created to collect biomolecular complexes from the PDB database \citep{burley2020rcsb}, with experimental binding affinity data curated from original reference papers. The latest release of PDBbind (version 2020) consists of over 23,000 biomolecular complexes, whose majority are protein-ligand complexes (19,443) and protein-protein complexes (2,852), and the remaining part are mainly protein-nucleic acid and nucleic acid-ligand complexes.

\subsubsection{Methods}

It is often critical to access the target structure before estimating the affinity of candidate molecules, since the affinity is jointly determined by the interaction between molecule and target. Depending on the availability of known target structures, affinity prediction methods can be roughly divided into two categories: ligand-based and structure-based.

\paragraph{Ligand-based affinity prediction (LBAP).} Based on the hypothesis that structurally analogous compounds tend to have similar biological activities \citep{Johnson1990}, ligand-based affinity prediction methods are developed. The ultimate goal is to identify promising compounds from a large candidate library, based on their similarities to known active compounds for an interested target. Several approaches are proposed to filter compounds based on chemical similarity measurements, {e.g.,} Tanimoto coefficients \citep{kim2008assessment} and similarity ensemble approach (SEA) \citep{Keiser2007}. Such methods heavily rely on hand-crafted or learnt compound representations, describing various properties including molecule weight, geometry, volume, surface areas, ring content, etc. On the other hand, quantitative structure-activity relationship (QSAR) based approaches attempt to explicitly formulate the relationship between structural properties of chemical compounds and their biological activities \citep{Kwon2019}. Various machine learning techniques have been cooperated with QSAR-based affinity prediction, including linear regression \citep{luco1997qsar}, random forest \citep{svetnik2003random}, support vector machine \citep{zakharov2016qsar}, and neural networks \citep{burden1999robust, pradeep2016ensemble}. Particularly, multi-task neural networks \citep{dahl2014multitask} alleviate the over-fitting issue by optimizing over multiple bioassasys simultaneously, and was adopted to achieve the best performance in the Merck Molecular Activity Challenge\footnotemark{}.
\footnotetext{Merck Molecular Activity Challenge: \url{https://www.kaggle.com/c/MerckActivity}}

Despite the satisfying performance of ligand-based affinity prediction approaches in certain scenarios, they do not take target structures in consideration. However, the interaction between target and molecule is indeed essential in accurately predicting the binding affinity, which leads more and more researches to focus on structure-based affinity prediction.

\paragraph{Structure-based affinity prediction (SBAP).} In contrast to ligand-based approaches, structure-based methods \citep{lim2021review} usually take structures of protein targets and/or protein-ligand complexes as inputs for affinity prediction. Some work \citep{wallach2015atomnet, li2021structure} predicts the binding affinity from experimentally determined protein-ligand co-crystal structures, but such data is highly expensive and time-consuming to obtain in practice. Others turn to computation-based docking routines \citep{trott2010autodock, koes2013lessons, mcnutt2021gnina, bao2021deepbsp} to estimate protein-ligand complex structures through sampling and ranking, and then formulate the structure-affinity relationship via various models.

\cite{Ballester2010} propose the RF-Score approach to use random forest to implicitly capture binding effects based on a series of carefully designed hand-crafted features. 3D convolutional neural networks are adopted in \citep{stepniewska2018development, jimenez2018k}, where protein-ligand complex structures are discretized into 3D voxels and then fed into the model for affinity prediction. However, such discretization fails to capture the 3D rotational and translational invariance of 3D structures, and thus relies on heavy data augmentation to overcome such limitations. Graph neural networks are adopted in \cite{Jiang2021} to simultaneously formulate the intra-molecular and inter-molecular interactions, where nodes correspond to ligand/protein atoms, and edges are defined by both covalent and non-covalent linkages.

\subsubsection{Discussions}

For most machine learning based approaches, it is usually desirable that the data distribution of training and evaluation subsets are as close as possible. However, this often does not hold true for affinity prediction tasks, {e.g.,} the scaffold of small molecules and/or family of protein targets encountered during inference may be unseen throughout the model training process. Simply dividing the database into training and evaluation subsets on a per target-molecule basis may lead to over-optimistic performance, which is unrealistic for real-world applications.

Nonetheless, most databases for experimental binding affinities do not provide an official data split or its generation pipeline for model training and evaluation. To make things even more complicated, binding affinity annotations could be highly noisy, due to different experimental settings, affinity measurements, and confidence scores. Researchers need to manually pre-process raw data entries and re-organize them into the standard format, which is not only laborious and burdensome, but also unfavorable for a fair comparison against existing baselines. Therefore, we propose \drugood{} as a highly customizable curator for OOD datasets with noisy labels explicitly considered, so as to promote more efficient development of affinity prediction approaches in \aidd.

\subsection{General OOD Databases}

The out-of-distribution issue has attracted an ever-growing research interest in recent years, due to its importance in improving the generalization ability in real-world applications. Several databases have been constructed with great emphasis placed on the out-of-distribution generalization performance, mainly consist of computer vision and natural language processing tasks.

In \cite{WILDS}, the WILDS benchmark is proposed to reflect various levels of distribution shifts that may occur in real-world scenarios. It considers two common types of distribution shifts: domain generalization and sub-population shift. A total of 10 datasets are included, covering shifts across cameras for wildlife monitoring, hospitals for tumor identification, users for product rating estimation, andƒ scaffolds for biochemical property prediction, etc. \cite{sagawa2021extending} further extend this database to include unlabeled data for unsupervised domain adaptation.

DomainBed \citep{gulrajani2020search} consists of 7 multi-domain image classification datasets, including Colored MNIST \citep{arjovsky2019invariant}, Rotated MNIST \citep{ghifary2015domain}, PACS \citep{li2017deeper}, VLCS \citep{fang2013unbiased}, Office-Home \citep{venkateswara2017deep}, Terra Incognita \citep{beery2018recognition}, and DomainNet \citep{peng2019moment}. Furthermore, authors point out the importance of model selection strategy in the domain generalization task, and conduct thorough benchmark tests over 9 baseline algorithms and 3 model selection criteria. As it turns out, empirical risk minimization (ERM) \citep{vapnik1999nature} with a careful implementation achieves state-of-the-art performance across all datasets, even when compared against various domain generalization algorithms.

\cite{ye2021ood} analyze the performance comparison between ERM and domain generalization algorithms on DomainBed, and point out that the distribution shift is composed of diversity shift and correlation shift, and existing domain generalization algorithms are only optimized towards one of them. They further propose additional datasets, of which WILDS-Camelyon17 \citep{WILDS} is dominated by diversity shift, and NICO \citep{he2021towards} and CelebA \citep{liu2015deep} are dominated by correlation shift.

As described above, general OOD databases are mostly built with image and text data, with one exception being the ODGB-MolPCBA dataset from WILDS \citep{WILDS}, which aims at predicting biochemical properties from molecular graphs. The distribution shift is mainly caused by disjoint molecular scaffolds between training and test subsets. This is indeed critical for accurate prediction of target-independent properties, but is still insufficient for affinity prediction where the target information should be explicitly exploited. TDC \citep{huang2021therapeutics} as another concurrent AIDD benchmark, offers SBAP datasets collated from BindingDB with a temporal split by patent year between 2013-2021, which is still limited in scope for drug OOD problems. 
In contrast, \drugood covers comprehensive sources of out-of-distribution in affinity prediction,
and provides a dataset curator for highly customizable generation of OOD datasets. Additionally, noisy annotations are taken into consideration, so that algorithms can be evaluated in a more realistic setting, which further bridges the gap between researches and pharmaceutical applications.

\subsection{Benchmark Scheme}

In additional to an automated dataset curator in \drugood{}, we also provide rigorous benchmark tests over state-of-the-art OOD algorithms, with graph neural networks and BERT-like models used for representation learning from structural and sequential data. Next, we briefly review these OOD and representation learning algorithms, while more detailed descriptions can be found in  \cref{sec_benchmark_sota_ood_alg}.

\subsubsection{General Methods for OOD and Noisy Labels}

The out-of-distribution and noisy label issues have been extensively studied in the machine learning community, due to their importance in improving the generalization ability and robustness. Here, we summarize recent progress in these two areas respectively, with a few overlaps since some approaches are proposed to jointly tackle these two issues in one shot.

\paragraph{Methods for OOD.} To improve the model generalization ability over out-of-distribution test samples, some work focuses on aligning feature representations across different domains. The minimization of feature discrepancy can be conducted over various distance metrics, including second-order statistics \citep{sun2016deep}, maximum mean discrepancy \citep{tzeng2014deep} and Wasserstein distance \citep{zhou2021domain}, or measured by adversarial networks \citep{ganin2016domain}. Others apply data augmentation to generate new samples or domains to promote the consistency of feature representations, such as Mixup across existing domains \citep{xu2020adversarial, yan2020improve}, or in an adversarial manner \citep{zhao2020maximum, qiao2020learning}.

With the label distribution further taken into consideration, recent work aims at enhancing the correlation between domain-invariant representations and labels. For instance, invariant risk minimization \citep{arjovsky2019invariant} seeks for a data representation, so that the optimal classifier trained on top of this representation matches for all domains. Additional regularization terms are proposed to align gradients across domains \citep{koyama2021invariance}, reduce the variance of risks of all domains \citep{krueger2021distribution}, or smooth inter-domain interpolation paths \citep{chuang2021fair}.

\paragraph{Methods for noisy labels.} There is a rich body of literature trying to combat with the label-noise issue, starting from the seminal work \citep{angluin1988learning} for traditional statistical learning to recent work for deep learning \citep{han2018co, han2019deep, song2020learning}. Generally speaking, previous methods attempt to handle noisy labels mainly from three aspects \citep{han2020survey}: training data correction \citep{van2015learning, van2017theory}, objective function design \citep{azadi2015auxiliary, wang2017robust}, and optimization algorithm design \citep{jiang2018mentornet, han2018co}.

From the training data perspective, prior work \citep{van2017theory} firstly estimates the noise transition matrix, which characterizes the relationship between clean and noisy labels, and then employs the estimated matrix to correct noisy training labels. Typical work in this line includes using an adaptation layer to model the noise transition matrix \citep{sukhbaatar2014training}, label smoothing \citep{lukasik2020does}, and human-in-the-loop estimation \citep{han2018masking}.

Other work turns to the design of objective functions, which aims at introducing specific regularization into the original empirical loss function to mitigate the label noise effect. In \cite{azadi2015auxiliary}, authors propose a group sparse regularizer on the response of image classifier to force weights of irrelevant or noisy groups towards zero. \cite{zhang2017Mixup} introduce Mixup as an implicit regularizer, which constructs virtual training samples by linear interpolation between pairs of samples and their labels. Such approach not only regularizes the model to factor simple linear behavior, but also alleviates the memorization of corrupted labels. Similar work following this idea includes label ensemble \citep{laine2016temporal} and importance re-weighting \citep{liu2015classification}.

From the view of optimization algorithm design, some work \citep{han2018co,li2020dividemix} introduces novel optimization algorithms or schedule strategies to solve the label noise issue. One common approach in this line is to train a single neural network via small loss tricks \citep{ren2018learning, jiang2018mentornet}. Besides, other work proposes to co-train two neural networks via small-loss tricks, including co-teaching \citep{han2018co} and co-teaching+ \citep{yu2019does}.

\subsubsection{Representation Learning for Affinity Prediction}

As a fundamental building block of ligand- and structure-based affinity prediction, machine learning models for encoding molecules and proteins have attracted a lot of attention in the research community. Based on the input data and model type, existing studies can be divided into following three categories.

\paragraph{Hand-crafted feature based backbones.} Many early studies enlist domain experts to design features from molecules and proteins which contain rich biological and chemical knowledge, such as Morgan fingerprints \citep{morgan1965generation}, circular fingerprints \citep{glen2006circular}, and extended-connectivity fingerprint \citep{rogers2010extended}. Then, these hand-crafted features are fed to machine learning models to produce meaningful embeddings for downstream tasks. Typical models include logistic regression \citep{kleinbaum2002logistic}, random forest \citep{breiman2001random}, influence relevance voting \citep{swamidass2009influence}, and neural networks \citep{ramsundar2015massively}.

\paragraph{Sequence-based backbones.} Since both molecules and proteins have their own sequential formats, SMILES \citep{weininger1988smiles} and amino-acid sequence \citep{sanger1952arrangement}, it is reasonable to utilize models which can naively deal with the sequential data, including 1D-CNN \citep{hirohara2018convolutional}, RNN \citep{goh2017smiles2vec}, and BERT \citep{wang2019smiles, brandes2021proteinbert}. Specifically, some studies \citep{Xu2017Seq2seqFA, min2021pre} introduce self-supervised techniques from natural language processing to generate high-quality embeddings. 
However, 1D linearization of molecular structure highly depends on the traverse order of molecular graphs, which means that two atoms that are close in the sequence may be far apart and thus uncorrelated in the actual 2D/3D structures (e.g., the two oxygen atoms in "CC(CCCCCCO)O"), therefore hinders language models to learn effective representations that rely heavily on the relative position of tokens \citep{wang2021chemical}.

\paragraph{Graph-based backbones.} To eliminate the information loss in the sequence-based models, recently, some studies start to explore the more complex graph-based representation of molecules and proteins and utilize graph neural networks (GNNs) to produce embeddings which encode the information of chemical structures, such as GIN \citep{xu2018powerful}, GCN \citep{kipf2016semi}, and GAT \citep{velivckovic2017graph}. Other than directly applying the vanilla GNN model as the backbone, several studies try to incorporate the domain knowledge in the model design, such as ATi-FPGNN \citep{xiong2019pushing}, NF \citep{duvenaud2015convolutional}, Weave \citep{kearnes2016molecular}, MGCN \citep{lu2019molecular}, MV-GNN \citep{ma2020multi}, CMPNN \citep{song2020communicative}, MPNN \citep{gilmer2017neural}, and DMPNN \citep{yang2019analyzing}. Other than standard GNN models, several studies exploit Transformer-like GNN models to enhance the expressive power for training with a large number of molecules and proteins, such as GTransformer \citep{rong2020self} and Graphormer \citep{ying2021transformers}.

\section{Automated Dataset Curator with Real-world Domain and Noise Annotations} 
\label{sec_curator}

We construct all the datasets based  on \chembl \citep{mendez2019chembl}, which is a large-scale, open-access drug discovery database that aims to capture medicinal chemistry data and knowledge across the pharmaceutical research and development process. We use the latest release in the SQLite format:  \chembl 29\footnote{Download link:  \url{http://ftp.ebi.ac.uk/pub/databases/chembl/ChEMBLdb/releases/chembl_29/chembl_29_sqlite.tar.gz}}.
Moreover, we consider the setting of OOD and different noise levels, which is an inevitable problem when the machine learning model is applied to the drug development process.
For example, when predicting SBAP bioactivity in practice, the target protein used in the model inference could be very different from that in the training set and even does not belong to the same protein family. The real-world domain gap will invoke challenges to the accuracy of the model.
On the other hand, the data used in the wild often have various kinds of noise, e.g. activities measured through experiments often have different confidence levels and different ``cut-off'' noise. Therefore, it is necessary to construct data sets with varying levels of noise in order to better align with the real scenarios.

In the process of drug development, many predictive tasks are involved. Here, we consider  two crucial  tasks from computer aided drug discovery \citep{Sliwoski2013}: ligand based affinity  prediction (LBAP) and structure based affinity prediction (SBAP). 
In LBAP, we follow the common practice and do not involve any protein target  information, which is usually used in the activity prediction for one specific protein target.
In SBAP, we consider both the  target  and drug information to predict the binding activity of ligands, aiming to develop models that can generalize across different protein targets.

\begin{figure}[!t]
	\centering  
	\includegraphics[width=1\textwidth]{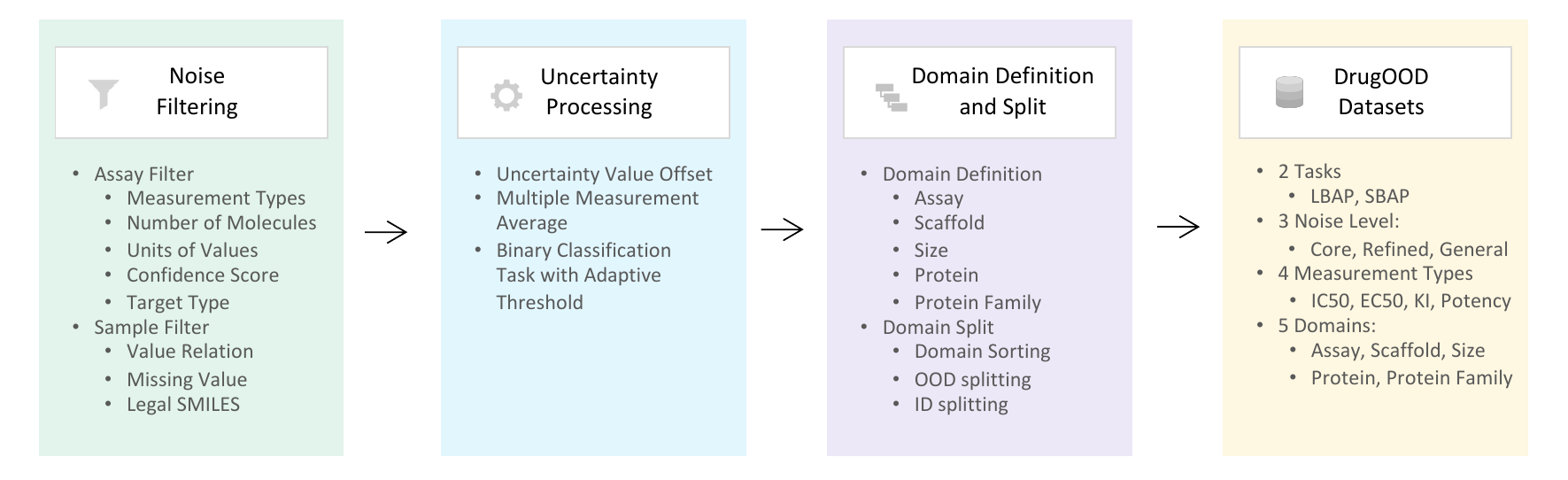}
	\caption{Overview of the automated dataset curator. We mainly implement three major steps based on the ChEMBL data source: noise filtering, uncertainty processing, and domain splitting. We have built-in 96 configuration files to generate the realized  datasets with the configuration of two tasks, three noise levels, four measurement types, and five domains.}
	\label{fig:intro:curator}
\end{figure}

\subsection{Filtering Data with Different Noise Levels}
\chembl contains the experimental results reported in previously published papers, sorted by ``assays''. However, the activity values for different assays and even different molecules in the same assay may have different accuracy and confidence levels. For example, assays that record the results of high-throughput screens (HTS) usually have less accurate activity values. The data with different accuracy constitute different noise annotations. 
Therefore, we set up filters with different stringency to screen for datasets with varying noise levels. 

Our filters consist assay filters and sample filters, which aim to screen for the required assays and samples.  Specifically, we have built-in 5 assay filters and 3 sample filters in advance. The details are shown as follows.

\begin{snugshade}
\begin{itemize}
\item[$\blacktriangleright$] \textbf{Assay Filter}
\begin{itemize}
    \item[$\bullet$]  \textbf{Measurement Type:} Filtering out the assays with required measurement types, e.g. IC50, EC50. Due to the big difference in meaning between measurement types, and its difficulty to merge them, we generate different datasets for each measurement types.
    \item[$\bullet$] \textbf{Number of Molecules:} Assay noise is strongly related to the number of molecules in an assay. For example, large assays are often derived from high-throughput screens that are very noisy.  Hence we set different molecules number limits for different noise levels.

    \item[$\bullet$] \textbf{Units of Values:} The units of activity values recorded in \chembl are chaotic, e.g. nM, \%, None. The conversion between some units is easy, such as nM and $\mu$M. But most of them cannot be converted between each other.
    \item[$\bullet$] \textbf{Confidence Score:} Due to the complex settings of different experimental assays, sometimes it is not certain that the target that interacts with the compound is our designated target. Confidence score of an assay indicating how accurately the assigned target(s) represent(s) the actually assay target.
    \item[$\bullet$] \textbf{Target Type:} There are dozens of target types in \chembl, from 'single protein' to 'organism'. Different target types have different confidence levels. For SBAP tasks, the target type of a 'single protein' is more reliable than others.

\end{itemize}

\item[$\blacktriangleright$] \textbf{Sample Filter}
\begin{itemize}
    \item[$\bullet$] \textbf{Value Relation:} 
    In many cases, the precise activity values of certain molecules could not be obtained, but given a rough range.
    \chembl records qualitative values with different relationships, e.g. '=', '>', '<', '>>'. Obviously, '=' means the value is  accurate, and the others  means the value is not accurate.
    \item[$\bullet$] \textbf{Missing Value:} Filtering samples with any missing value.
    \item[$\bullet$] \textbf{Legal SMILES:} Filtering samples with illegal molecules.
\end{itemize}

\end{itemize}
\end{snugshade}

The configurations of each filter for three different noise levels are shown in the \cref{tab:detail_config_noise}.
One can see that the three noise levels are now annotated jointly by ``Confidence Score'', ``Value Relation'', ``Number of Molecules'' and ``Target Type'', which are shown in blue color.
In  \cref{fig:overview}, we display some examples of the SBAP dataset, demonstrating different kinds of noise levels.

\setlength{\tabcolsep}{10pt}
\renewcommand{\arraystretch}{1}
\begin{table}[]
\caption{The filter configurations for three noise levels.  $\divideontimes$ means the user configuration will specify the measurement type. '---' denotes no restriction. \checkmark  indicates the conditions need to be met. "SP": single protein. "PC": protein complex. "PF": protein family. One can see that the three noise levels are now annotated jointly by “Confidence Score”, "Value Relation", "Number of Molecules" and "Target Type", which are shown in blue color.}   
\label{tab:detail_config_noise}
\small 
\centering
\begin{tabular}{c|cccc} \hline 
 & Name & Core & Refined & General \\
\hline
\multirow{5}{*}{\shortstack{Assay\\Filter}}  
& {Measurement Type} & $\divideontimes$ & $\divideontimes$  & $\divideontimes$ \\
& \textcolor{cite_color}{Number of Molecules} & $[50,3000]$ & $[32,5000]$ & $[10,5000]$ \\
& Units of Values  & $\{nM,uM\}$ & $\{nM,uM\}$ & $\{nM,uM\}$ \\
& \textcolor{cite_color}{Confidence Score} & $\geq 9$ &$\geq 3$ & --- \\
& \textcolor{cite_color}{Target Type}      & $\{SP\}$&$\{SP, PC, PF\}$ & ---\\
\hline
\multirow{3}{*}{\shortstack{Sample\\Filter}} 
& \textcolor{cite_color}{Value Relation} & $\{=,\sim \}$ & $\{=,\sim,\geq,\leq \}$ & $\{=,\sim,\geq,\leq, >, < \}$ \\
& Missing Value  & \checkmark & \checkmark & \checkmark \\
& Legal SMILES & \checkmark & \checkmark & \checkmark \\
\hline
\end{tabular}
\end{table}

\subsection{Processing Uncertainty and Multiple Measurements}

As mentioned above, \chembl records many activity values in an uncertainty way, and they were reported as above or below the highest or lowest concentration tested. Here, we follow the practice in pQSAR 2.0~\citep{Profile-QSAR-2.0} and offset them by 10-fold. 
Meanwhile, since the same molecule may be reported in different sources, the same molecule may also appear in multiple assays in \chembl. We call this phenomenon ``multiple measurements''. 
Following the common practice~\citep{mean_for_multi_measurement}, we average all the multiple measurements. For LBAP task, we average all the activity values for the same molecule before domain split. While for SBAP task, when the molecule and target pair is the same, we average their activity values.

\subsection{Binary Classification Task with Adaptive Threshold}
While \chembl record activity values as floating numbers, benchmarking OOD tasks as regression tasks is known to be extremely hard because  various noises such as uncertainty measurements. 
Additionally, in the process of drug development, practitioners habitually consider whether a compound is active/inactive. Therefore, using a binary classification task is more robust and good enough to make decisions in the drug development.
However, in practice the threshold for binary classification depends on the specific circumstances of the drug development project. 
Here, we choose to use an adaptive threshold method that can adapt to a wider range of situations. 
In particular, the median value of all compounds in the generated dataset defines the threshold, but the range of allowed thresholds are fixed to be $4 \leq pValue \leq 6$, where $pValue = -\log_{10}(Activity~Value)$. 
If the median is outside this range, a fixed threshold $pValue = 5$ is applied, which follows the common practice~\citep{Autothr,fsmol} in drug discovery.
In this way, we can try our best to keep the dataset balanced while making the generated tasks meaningful.

\subsection{Domain Definition and Split}
\label{subsec_domain_split_id}

\subsubsection{Domain Definition}
As mentioned before, distributional shift is a common phenomenon in the drug development process. In order to make our benchmark more in line with the needs of drug discovery and development, we consider the OOD setting in the benchmark. 
In the process of drug research and development, when predicting the bioactivities of small molecules, we may encounter very different molecular scaffolds, sizes and so on from the model training set.
These differences may also be reflected in the target in the SBAP task.
Hence, for LBAP task, we consider the following three domains: assay, scaffold and molecule size. For SBAP tasks, in addition to the three domains mentioned above, we also consider two additional target-specific domains: protein and protein family.
The user can also easily customize the domain through the configuration file and generate the corresponding dataset.
The details of the five domains are  as follows.

\begin{snugshade}
\begin{itemize}
    \item[$\bullet$] \textbf{Assay:} Samples in the same assay are put into the same domain. Due to the great differences in different assay  environments, the activity values measured by different assays will have a large shift. At the same time, different assays have very different types of experiments and test targets.
    \item[$\bullet$] \textbf{Scaffold:} Samples with the same molecular scaffold belong to the same domain.  The molecular properties of different scaffolds are often quite different.
    \item[$\bullet$] \textbf{Size:} A domain consists of samples with the \emph{same} number of atoms. As a result, we can test the model's performance on molecules that are quite different.

    \item[$\bullet$] \textbf{Protein:} In SBAP task, samples with the same target protein are in the same domain, in order to test the model performance when meeting a never-seen-before protein target.
    \item[$\bullet$] \textbf{Protein Family:} In SBAP tasks, samples with targets from the same family of proteins are in the same domain.  Compared with protein domains, there are much less protein family domains albeit with greater differences among each other.
\end{itemize}
\end{snugshade}

\begin{figure}[!t]
	\centering 
	\includegraphics[width=0.85\textwidth]{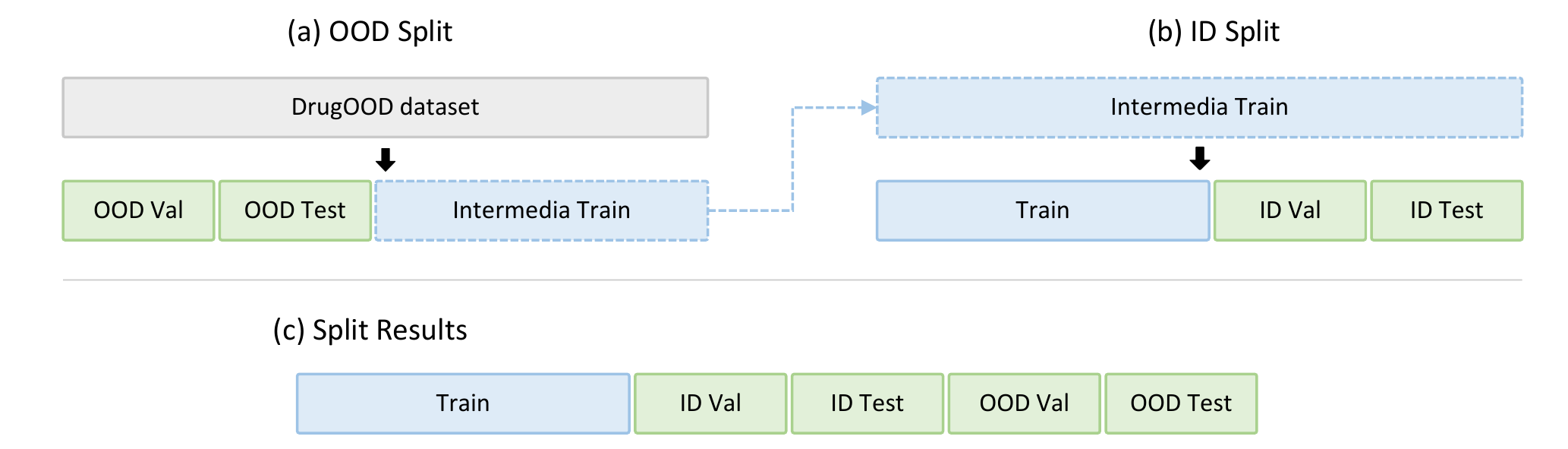}
	\caption{Illustration of our domain split process: (a) \textit{OOD Split:} Based on the sorted data, we sequentially split them into an intermediate training set, OOD validation, and testing set. 
	We try to keep the proportion of training, validation, and testing samples at around 6:2:2. (b) \textit{ID Split:} From the intermediate training set, we further separate the final train, ID validation, and ID test sets. (c) We get training, ID validation, ID test, OOD validation, and OOD test sets after executing the above steps.}
	\label{fig:intro:split}
\end{figure}

\subsubsection{Domain Split}
Based on the generated domains, we need to split them into training, OOD validation and OOD testing set. 
Our goal is to make the domain shifts between the training set and the OOD validation/testing set as significant as possible.
This leads to the question of how to measure the differences between different domains and how to sort them into the training set and validation/testing set.
Here, we design a general pipeline, that is, firstly generate domain descriptor for each domain, and then sort the domains with descriptors. Then the sorted domains are sequentially divided into training set, OOD validation and testing set (\cref{fig:intro:split} (a)).
Meanwhile, the number of domains in different splits are controlled by the number of total samples in each splits, and the proportion of  sample numbers is trying to be kept at 6:2:2 for training, validation and testing set.

In the \drugood framework, we have built-in two domain descriptor generation methods as follows.
\begin{snugshade}
\begin{itemize}
    \item[$\bullet$] \textbf{Domain Capacity:} The domain descriptor is the number of samples in this domain. In practice, we found that the number of samples in a domain can well represent the characteristics of a domain. For example, an assay with 5000 molecules is usually different from an assay with 10 molecules in assay type. For SBAP, a protein family that contains 1000 different proteins is very likely to be different from the protein family that contains only 10 proteins. The descriptor is applied to the domains of assay, protein and protein family.
    \item[$\bullet$] \textbf{Molecular Size:} For the size domain, the size itself is already a good domain descriptor and can be applied to sort the domains, which we found to effectively increase the generalization gap in practice. The descriptor is applied to the domain of size and scaffold.
\end{itemize}
\end{snugshade}

\paragraph{Generating in-distribution data subsets.}
After splitting the training, OOD validation and OOD test sets, we split the ID validation and ID testing sets from the resultant training set  (\cref{fig:intro:split} (b)). We follow similar settings as of WILDS ~\citep{WILDS}, the ID validation and testing sets are merged from the randomly selected samples of each domain in the training set. 
The ratio of the number of samples in the OOD and ID validation/test sets can be easily modified through the configuration files. After this process, we get the final training, ID validation, ID test, OOD validation, OOD test sets (\cref{fig:intro:split} (c)).  

\setlength{\tabcolsep}{30pt}
\renewcommand{\arraystretch}{1}
\begin{table}[]
\caption{Statistical information of datasets in LBAP tasks under the IC50 measurement type. $Domain^{\#}$ represents the number of domains, and $Sample^{\#}$ represents the number of data points.}
\label{tab:data_statistics_lbap_task}
\small 
\centering
\begin{tabular}{lcc} \hline 
Data subset & $Domain^{\#}$  &$Sample^{\#}$ \\ \hline\hline
\drugood-lbap-core-ic50-assay &1,324 &95,236 \\
\drugood-lbap-refined-ic50-assay &5,612 &266,522 \\
\drugood-lbap-general-ic50-assay &29,938 &568,556 \\ \hline
\drugood-lbap-core-ic50-scaffold &34,807 &95,236 \\
\drugood-lbap-refined-ic50-scaffold &90,888 &266,522 \\
\drugood-lbap-general-ic50-scaffold &186,875 &568,556 \\ \hline
\drugood-lbap-core-ic50-size &251 &95,236 \\ 
\drugood-lbap-refined-ic50-size &288 &266,522 \\
\drugood-lbap-general-ic50-size &337 &568,556 \\ \hline
\end{tabular}
\end{table}

\subsection{Overview of the Dataset Curation Pipeline}

The overview of the datasets curation pipeline is shown in \cref{fig:intro:curator}. It mainly includes four major steps: filtering data with different noise levels, processing uncertainty and multi-measurements,  binary classification task with adaptive threshold and domain split. 
We have built-in 96 configuration files to generate different datasets with the configuration of 2 tasks, 3 noise levels, 4 measurement types, and 5 domains. With our \drugood dataset curation, the user can easily obtain the required datasets through customizing the configuration files.

\begin{figure}[!t]
\centering
\scalebarimg{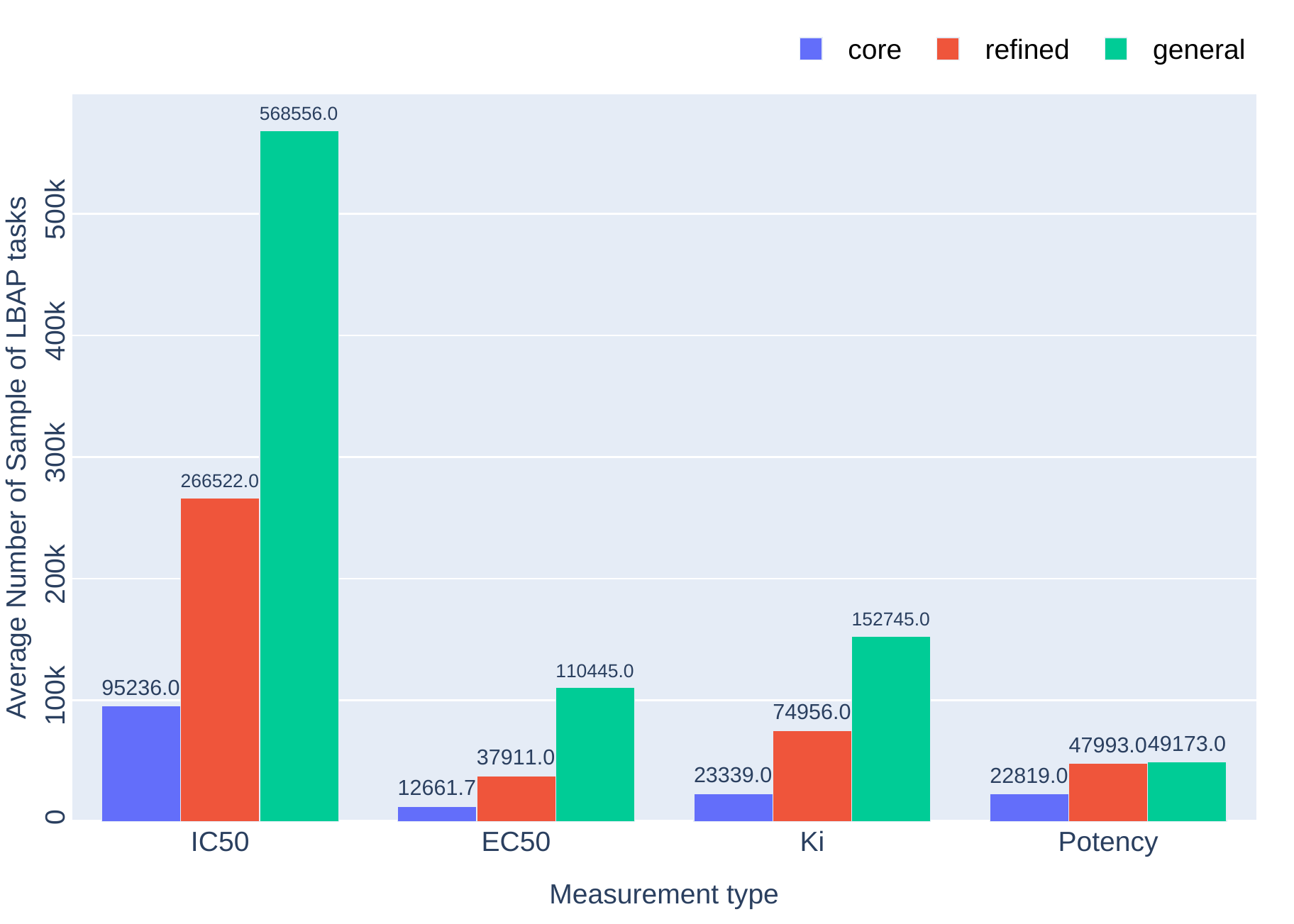}{0.4}{(a)}
\scalebarimg{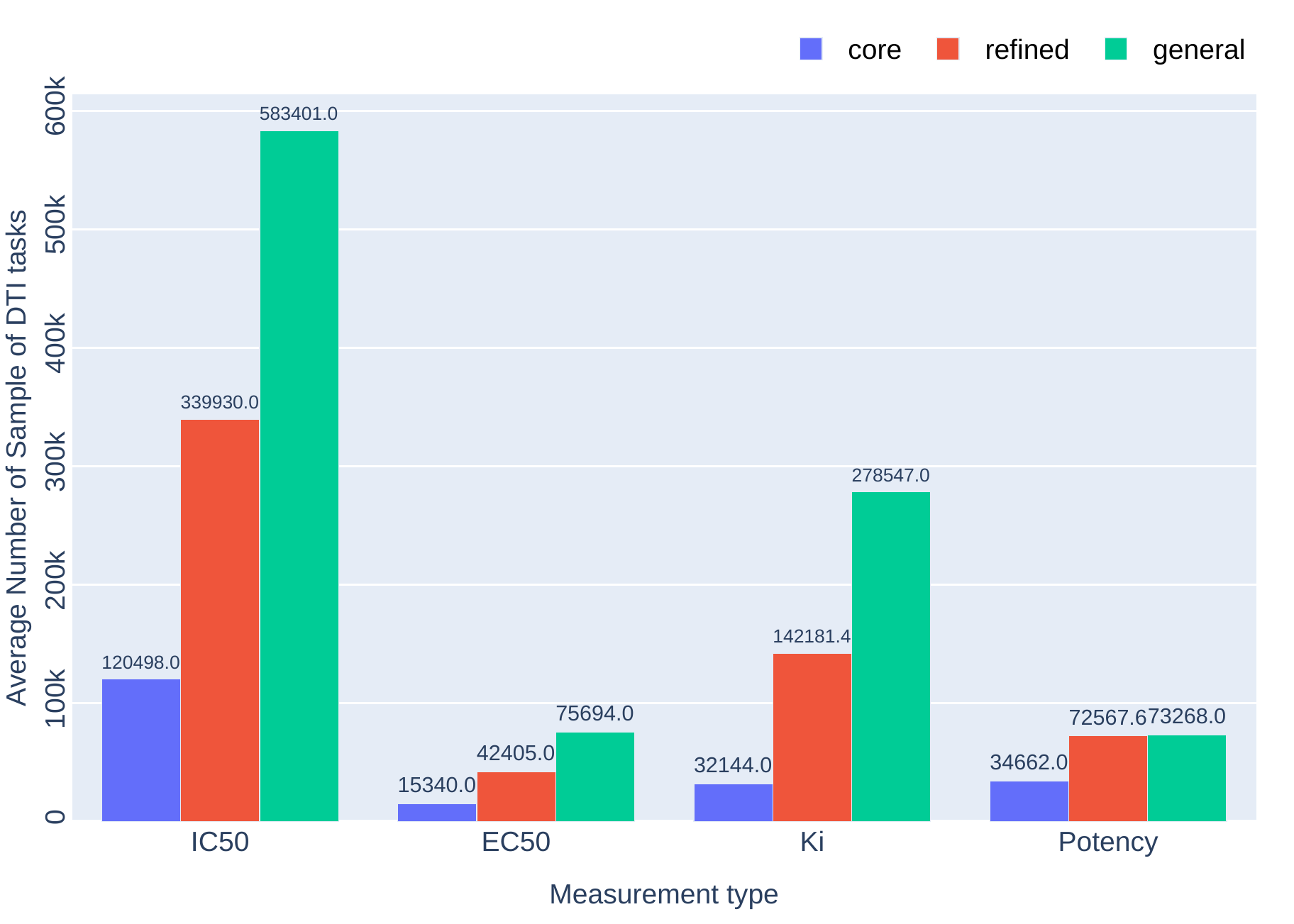}{0.4}{(b)}
\caption{Statistics of samples for (a) LBAP and (b) SBAP tasks.
Each column represents the average of all domain sample sizes under a certain noise level and measurement type.
}
\label{fig:statistics_samples}
\end{figure}

Here, we show some statistics of datasets generated by our built-in configuration files.
\cref{tab:data_statistics_lbap_task} and \cref{tab:data_statistics_sbap_task} show the statistics of domains and samples in LBAP task and SBAP task under IC50 measurement type, respectively. 
We can see that as the number of samples increases, the noise level
also increases.
Meanwhile, for the same noise level, there are huge differences in the number of domains generated by different domain split methods, which  will challenge the applicability of OOD algorithm in different domain numbers.
In order to show the comparison of data volume under different measurement types, we count the samples of different measurement types under different noise levels, as shown in ~\cref{fig:statistics_samples}.
As we can see, the number of samples varies greatly under different measurement types in \chembl. Meanwhile, different measurement types may also bring different noise levels. Our curation can generate specific measurement types of datasets according to the needs of specific drug development scenarios.
More statistical information of the \drugood datasets under different setting are summarized in \cref{tab:data_statistics_lbap_task:b} and \cref{tab:data_statistics_sbap_task:b}.

\setlength{\tabcolsep}{35pt}
\renewcommand{\arraystretch}{1}
\begin{table}[]
\caption{Statistical information of  some \drugood datasets. $Domain^{\#}$ represents the number of domains, and $Sample^{\#}$ represents the number of data points.}
\scriptsize
\centering
\begin{tabular}{lcc} \hline 
Data subset & $Domain^{\#}$  &$Sample^{\#}$ \\ \hline\hline

\drugood-sbap-core-ic50-assay  & 1,455 & 120,498 \\
\drugood-sbap-refined-ic50-assay & 6,409 & 339,930 \\
\drugood-sbap-general-ic50-assay  & 23,868 & 583,401 \\ \hline
\drugood-sbap-core-ic50-scaffold & 34,807 & 120,498 \\
\drugood-sbap-refined-ic50-scaffold & 89,627 & 339,930 \\
\drugood-sbap-general-ic50-scaffold & 145,768 & 583,401 \\ \hline
\drugood-sbap-core-ic50-size & 251 & 120,498 \\ 
\drugood-sbap-refined-ic50-size & 288 & 339,930 \\
\drugood-sbap-general-ic50-size & 327 & 583,401 \\ \hline
\drugood-sbap-core-ic50-protein & 677 & 120,498 \\ 
\drugood-sbap-refined-ic50-protein & 1,483 & 339,930 \\
\drugood-sbap-general-ic50-protein & 2,671 & 583,401 \\ \hline
\drugood-sbap-core-ic50-protein-family & 13 & 120,498\\
\drugood-sbap-refined-ic50-protein-family & 15 & 339,930 \\ 
\drugood-sbap-general-ic50-protein-family& 15 & 583,401 \\ 
\hline
\end{tabular}
\label{tab:data_statistics_sbap_task}
\end{table}

\section{Benchmarking State-of-the-art OOD Algorithms}
\label{sec_benchmark_sota_ood_alg}
Our benchmark implements and evaluates algorithms from various perspectives, including \emph{architecture design} and \emph{domain generalization} methods, to cover a wide range of approaches to addressing the distribution shift problem. We believe this is the first paper to comprehensively evaluate a large set of approaches under various settings for the \drugood problem.

\subsection{Architecture Design}
It's known that the expressive power of the model is largely depend on the network architecture. How to design the network architecture for the better ability to fit the target function and robustness to the noise is a popular area of out-of-distribution data problem.
Based on the DGL-LifeSci package \citep{li2021dgl}, we benchmark and evaluate following graph-based backbones:
GIN \citep{xu2018powerful}, GCN \citep{kipf2016semi}, GAT \citep{velivckovic2017graph}, SchNet \citep{schutt2017schnet}, Weave \citep{kearnes2016molecular}, MGCN \citep{lu2019molecular}, ATi-FPGNN \citep{xiong2019pushing}, NF \citep{duvenaud2015convolutional} and  GTransformer \citep{rong2020self}.
For the Sequence based input, we adopt the BERT \citep{devlin2018bert} and Protein-BERT \citep{brandes2021proteinbert} as feature extractors.
The backbones are extended to regression and classification tasks by a readout function and an MLP layer.

We use a standard model structure for each type of data: GIN \citep{kipf2016semi} for molecular graphs and BERT \citep{devlin2018bert} for protein amino acid sequences.
In addition, the other models mentioned above were also used for some of the datasets in measuring the effect of model structure on generalization ability.
Following the model selection strategy in \citep{WILDS},  we use a distinct OOD validation set for model early stopping and hyper-parameter tuning.
The OOD validation set is drawn from a similar distribution of  the training set,  which distant from the OOD test set. 
For example, in the assay-based datasets, the training, validation, testing each consists of molecules from distinct sets of the assay environment.
We detail the experimental protocol in \cref{sec_exps}.

\setlength{\tabcolsep}{20pt}
\renewcommand{\arraystretch}{1}
\begin{table}[!t]
\scriptsize
\centering
\caption{The in-distribution (ID) vs out of distribution (OOD) of datasets with measurement type of IC50 trained with empirical risk minimization. The ID test datasets are drawn from the training data with the same distribution, and the OOD test data are distinct from the training data, which are described in \cref{sec_curator}. We adopt the Area under the ROC Curve (AUROC) to estimate model performance; the higher score is better. In all tables in this paper, we report in parentheses the standard deviation of 3 replications, which measures the variability among replications. All datasets show performance drops due to distribution shift, with substantially better ID performance than OOD performance. More experimental results under different setting are shown in \cref{tab:exp:lbap:erm_drop:c} in Appendix.}
\begin{tabular}{lccc} \hline 
Dataset & In-dist & Out-of-Dist & Gap \\ \hline\hline
\drugood-lbap-core-ic50-assay&89.62 (2.04) &71.98 (0.29) &17.64\\
\drugood-lbap-core-ic50-scaffold&87.15 (0.48) &69.54 (0.52) &17.60\\
\drugood-lbap-core-ic50-size&92.35 (0.15) &67.48 (0.47) &24.87\\
\drugood-lbap-refined-ic50-assay&89.25 (0.64) &72.70 (0.00) &16.55\\
\drugood-lbap-refined-ic50-scaffold&86.23 (0.08) &70.45 (0.54) &15.78\\
\drugood-lbap-refined-ic50-size&91.31 (0.07) &68.74 (0.37) &22.58\\
\drugood-lbap-general-ic50-assay&85.19 (1.15) &69.88 (0.13) &15.32\\
\drugood-lbap-general-ic50-scaffold&85.15 (0.24) &67.55 (0.09) &17.60\\
\drugood-lbap-general-ic50-size&89.77 (0.08) &66.05 (0.32) &23.72\\ \hline

\drugood-sbap-core-ic50-protein&90.71 (0.29) &68.87 (0.53) &21.84\\
\drugood-sbap-core-ic50-protein-family&89.88 (1.44) &72.20 (0.14) &17.68\\
\drugood-sbap-refined-ic50-protein&86.87 (1.41) &69.51 (0.30) &17.36\\
\drugood-sbap-refined-ic50-protein-family&86.44 (3.07) &70.61 (0.42) &15.82\\
\drugood-sbap-general-ic50-protein&85.34 (1.67) &68.48 (0.27) &16.86\\
\drugood-sbap-general-ic50-protein-family&79.18 (2.69) &68.60 (0.68) &10.58\\
\hline
\end{tabular}
\label{tab:exp:lbap:erm_drop:a}
\end{table}

\cref{tab:exp:lbap:erm_drop:a} shows that for each dataset with the measurement type of IC50, the OOD performance  is always significantly lower than the performance of the corresponding ID setting.

\subsection{Domain Generalization Algorithms}

In machine learning, models are commonly optimized by the empirical risk minimization (ERM), which trains the model to minimize the average training loss.
To improve the model robustness under the distribution shift, current methods tend to learn invariant representations that can generalize across domains.
There are two main directions: domain alignment and invariant predictors. 
Common approaches to domain alignment is to minimize the divergence of feature distributions from different
domains across distance metrics, such as maximum mean discrepancy \citep{tzeng2014deep, long2015learning} and adversarial loss \citep{ganin2016domain, li2018domain}, Wasserstein distance \citep{zhou2020domain}.
In addition, other conventional methods along this line of research are adopting data augmentation.
For example, Mixup \citep{zhang2017Mixup} augmentation proposes to construct additional virtual training data by convex combination of both samples and labels from the original datasets.
Follow-up works applied a similar idea to generate more domain and enhance consistency of features during training \citep{yue2019domain, zhou2020deep, xu2020adversarial, yan2020improve, shu2021open, wang2020heterogeneous, yao2022improving}, or synthesize unseen domain in an adversarial way to imitate the challenging test domains \citep{zhao2020maximum, qiao2020learning, volpi2018generalizing}.

\setlength{\tabcolsep}{1.6pt}
\renewcommand{\arraystretch}{1}
\begin{table}[!tp]\centering
\scriptsize
\caption{The out-of-distribution (OOD) performance of baseline models trained with different OOD algorithms on the \drugood-lbap-ic50 dataset.}
\begin{tabular}{l|ccc|ccc|ccc} \hline
\multirow{2}{*}{Algorithms} &\multicolumn{3}{c}{Assay} &\multicolumn{3}{|c|}{Scaffold} &\multicolumn{3}{c}{Size} \\
&core &refined &general &core &refined &general &core &refined &general \\ \hline \hline
ERM &71.98 (0.29) &69.54 (0.52) &67.48 (0.47) &69.88 (0.13) &67.55 (0.09) &66.05 (0.32) &72.70 (0.00) &70.45 (0.54) &68.74 (0.37) \\
IRM &69.22 (0.51) &64.94 (0.30) &57.00 (0.39) &69.21 (0.16) &63.78 (0.53) &60.52 (0.35) &71.68 (0.53) &67.97 (0.66) &59.99 (0.48) \\
DeepCoral &71.76 (0.60) &68.54 (0.01) &57.31 (0.44) &69.47 (0.13) &64.84 (0.53) &61.09 (0.23) &71.45 (0.09) &69.35 (0.73) &62.01 (0.38) \\
DANN &70.08 (0.65) &66.37 (0.20) &63.45 (0.18) &65.12 (0.94) &62.90 (1.02) &64.74 (0.27) &67.20 (0.68) &67.26 (0.95) &66.83 (0.08) \\
Mixup &71.34 (0.41) &69.29 (0.24) &67.73 (0.27) &69.87 (0.17) &67.64 (0.16) &66.01 (0.14) &72.13 (0.21) &70.68 (0.16) &68.69 (0.13) \\
GroupDro &71.54 (0.46) &66.67 (0.67) &60.90 (1.23) &69.36 (0.09) &64.91 (0.25) &61.41 (0.10) &71.40 (0.52) &69.08 (0.38) &62.15 (0.54) \\
\hline
\end{tabular}
\label{tab:exp:algorithm}
\end{table}
For learning invariant predictors, the core idea is to enhance the correlations between the invariant representation and the labels.
Representatively, invariant risk minimization (IRM) \citep{arjovsky2019invariant} penalizes the feature representation of each domain that has a different optimal linear classifier, intending to find a predictor that performs well in all domains.
Following up IRM, subsequent approaches propose powerful regularizers by penalizing variance of risk across domains \citep{krueger2021out}, by adjusting gradients across domains \citep{koyama2020out}, by smoothing interpolation paths across domains \citep{chuang2021fair}.
An alternative to IRM is to combat spurious domain correlation, a core challenge for the sub-population shift problem \citep{WILDS},
by directly optimizing the worst-group performance with Distributionally Robust Optimization \citep{sagawa2019distributionally, zhang2020coping, zhou2021examining}, generating additional samples around the minority groups \citep{goel2020model}, and re-weighting among groups with various size \citep{sagawa2020investigation}, or additional regulations \citep{chang2020invariant}.

We implement and evaluate the following representative OOD methods:
\begin{snugshade}
\begin{itemize}
    \item[$\bullet$] \textbf{ERM:} ERM optimizes the model by  minimizing the average empirical loss on observed training data. 
    \item[$\bullet$] \textbf{IRM:} IRM penalizes feature distributions for domains that have different optimum predictors.
    \item[$\bullet$] \textbf{DeepCoral:} DeepCoral penalizes differences in the means and covariances of the feature distributions (i.e., the distribution of last layer activations in a neural network) for each domain. Conceptually, DeepCoral is similar to other methods that encourage feature representations to have the same distribution across domains.
    \item[$\bullet$] \textbf{DANN:} Like IRM and DeepCoral, DANN encourages feature representations to be consistent across domains.
    \item[$\bullet$] \textbf{Mixup:} Mixup constructs additional virtual training examples from two examples drawn at random from the training data, which alleviates the effects of domain-related spurious information through data interpolation.
    \item[$\bullet$] \textbf{GroupDro:} GroupDro uses distributionally robust optimization methods to minimize worst-case losses, aiming to combat spurious correlations explicitly. 
\end{itemize}
\end{snugshade}

\cref{tab:exp:algorithm} summarizes the experimental results on the datasets with IC50 measurement type, showing that 
latest OOD algorithms exhibit \emph{no clear improvement} over the simple ERM algorithm.
There may be several reasons for this: 1). the molecular graph data are different from visual and textual input by nature, thus making it challenging to use conventional strategies directly; 2). these algorithms are usually designed for datasets that contain enough data per domain, so it is difficult to apply directly to datasets that have a large number of domains but few samples per domain.
There is a need for improved approaches to realistic \drugood problems, based on the results.
Additionally, current OOD research focus almost exclusively on the single-instance prediction tasks while overlook multi-instance prediction tasks, and how to better handle the distribution shift in multiple instance domains (e.g., molecule and protein inputs in the SBAP task) remains an open problem.
Lastly, while not explored in this paper, the large-scale realistic datasets always come with non-negligible inherent noise, both aleatoric and epistemic \citep{lazic2021quantifying}. And how to incorporate noise learning with OOD generalization to boost model's robustness and generalization in the meantime is an important research direction.

\section{Implementations and Package Usage}
\label{sec_package_usage}

\drugood develops a comprehensive benchmark for developing and evaluating OOD generalization algorithms.
Different with  other codebases, \drugood builds on the OpenMMLab project \citep{2019arXiv190607155C}, owning the following features: 

\begin{figure}[!tp]
\centering
\includegraphics[width=\columnwidth]{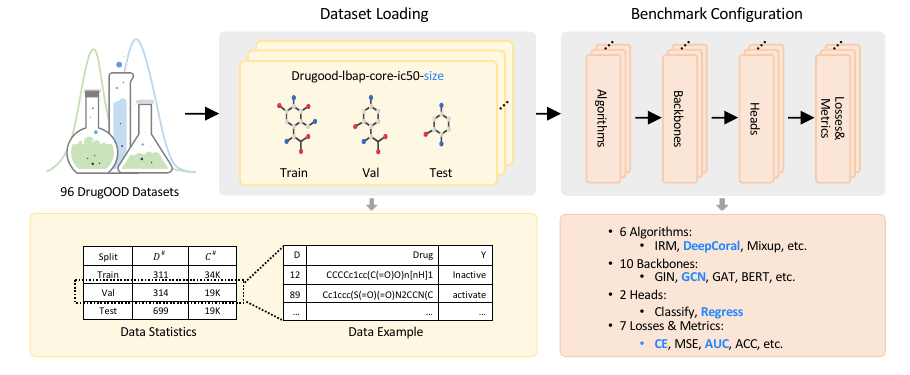}
\caption{Overview of the \drugood benchmark. \drugood conducts a comprehensive benchmark for developing and evaluating OOD generalization algorithms for \aidd. After loading any of the datasets generated by the data curator, users can flexibly combine different types of modules, including algorithms, backbones, etc., to develop OOD generalization algorithms in a flexible and disciplined manner.}
\label{fig:benchmark}
\end{figure}

\paragraph{Customizable dataset.}
\drugood supports various formats of data, providing related processing and converting tools. We provide 96 realized sub-datasets in advance. In addition to this, users can additionally specify additional conditions to easily customize new datasets from the original source dataset.

\paragraph{Modular design.}
Building on the design of OpenMMLab projects, we decompose the framework into different components and one can easily construct new OOD algorithms by combing these modules.

\paragraph{Support for multiple frameworks out of the box.}
\drugood codebase directly supports various  popular and contemporary OOD generalization algorithms, e.g. DeepCoral, IRM, DANN, Mixup etc.

With the above abstraction, our benchmark framework is illustrated in  \cref{fig:benchmark}. By simply creating some new components and assembling existing ones, the researchers can develop their approach efficiently.

\subsection{Dataset Curation}

The \drugood package provides a simple, standardized dataset curator based on the large-scale bioassay deposition website \chembl \citep{mendez2019chembl}, by proving a modified curation files, the researcher can easily re-configure the curation process. Specifically, we have provided 96 built-in configuration files for generating OOD dataset spanning various domains, noise annotations and measurement types.
The Listing \ref{lst:imp:curation} provides a simple example, which covers all of the steps of generating an OOD dataset from \chembl.
\begin{python}
# configure curation pipeline
curator = dict(
    task="lbap",
    chembl='../chembl_29_sqlite/chembl_29.db',
    save_dir=".data/lbap/lbap_core_EC50_assay",
    # noise filtering
    filter=dict(
        assay_filter=dict(
            measurement_type=["EC50"],
            assay_value_units=["nM", "uM"],
            molecules_number=[50, 3000],
            confidence_score=9
        ),
        sample_filter=dict(
            filter_none=True,
            smile_exist=True,
            smile_legal=True,
            value_relation=["=", "~"]
        )
    ),
    # uncertainty filtering
    uncertainty=dict(
        multiple_measurement_average=True,
        uncertainty_delta={'<': -1, '<=': -1, '>': 1, '>=': 1},
        binary_threshold=dict(
            lower_bound=4,
            upper_bound=6,
            fix_value=5)
    ),
    # domain split
    split=dict(
        domain=dict(
            domain_generate_field="assay_id",
            domain_name="assay",
            sort_func="domain_capacity"
            ),

        fractions=dict(
            train_fraction_ood=0.6,
            val_fraction_ood=0.2,
            IID_train_sample_fractions=0.6,
            IID_val_sample_fractions=0.2
        )

    ),
)
\end{python}
\begin{lstlisting}[frame=none,caption={Dataset curation example},captionpos=b,label=lst:imp:curation]
\end{lstlisting}

\subsection{Dataset Loading}
As shown in Listing \ref{lst:imp:dataset_loading}, \drugood provides a flexible and uniform interface for building data pipelines, allowing users to easily and quickly adjust the experimental data flow. Meanwhile, Standardized and automated evaluation of specified dataset partitions can be easily implemented by a few lines of code.

\begin{python}
dataset_type = 'MOL'
subset = 'lbap_ic50_core_assay'
data_prefix = "data/drugood"
# data prepossess pipeline
pipeline = [
    dict(type="SmileToGraph", keys=['input']),
    dict(type='ToTensor', keys=['gt_label']),
    dict(type='Collect', keys=['input', 'gt_label', 'meta'])
    ]
# data sampling strategy
sample_config = dict(
    uniform_over_groups=None,
    n_groups_per_batch=2,
    distinct_groups=True
    )
# data loading schedule
data = dict(
    samples_per_gpu=256,
    workers_per_gpu=0,
    train=dict(
        split="train",
        type=dataset_type,
        subset=subset,
        data_prefix=data_prefix,
        pipeline=pipeline,
        sample_mode="group",
        sample_config=sample_config
    ),
    val=dict(
        split="val",
        type=dataset_type,
        subset=subset,
        data_prefix=data_prefix,
        pipeline=pipeline,
        rule="greater", # model selection
        save_best="auc"
    ),
    test=dict(
        split="test",
        type=dataset_type,
        subset=subset,
        data_prefix=data_prefix,
        pipeline=pipeline)
    )
\end{python}
\begin{lstlisting}[frame=none,caption={Dataset loading example},captionpos=b,label=lst:imp:dataset_loading]
\end{lstlisting}

\subsection{Algorithm Configuration}
\drugood also supports popular and contemporary OOD generalization algorithm out of box. Users can easily configure different modules to construct and develop new OOD generalization algorithms effectively. We provide an example of building algorithm in few lines code, as shown in Listing \ref{lst:imp:algorithm}.

\begin{python}
# algorithm configuration
model = dict(
    type="ERM",
    # backbone 
    tasker=dict(
        type='Classifier',
        backbone=dict(
            type='GIN', # backbone setting
            num_node_emb_list=[39],
            num_edge_emb_list=[10],
            emb_dim=256,
            dropout=0.0,
        ),
        head=dict(
            type='LinearClsHead',
            num_classes=2,
            in_channels=256,
            loss=dict(type='CrossEntropyLoss'))
    ))
\end{python}
\begin{lstlisting}[frame=none,caption={Algorithm configuration example},captionpos=b,label=lst:imp:algorithm]
\end{lstlisting}

\section{Empirical Studies}
\label{sec_exps}

We present typical experimental results and corresponding analysis in this section.  More results and details are deferred to the Appendix. 

\subsection{Ligand Based Affinity Prediction (LBAP)}

Precisely predicting the affinity score of small molecules will greatly boost the process of drug discovery by reducing the needs of 
costly laboratory experiments.
However, the experimental data available for training such models is limited compared to the extremely diverse and combinatorially large universe of candidate molecules that we would want to make predictions on.
In this paper, we study the domain variation in experimental assay, molecules sizes, molecule sizes between training and test molecules.

\subsubsection{Setup}

\paragraph{Problem Definition.}
For the task of LBAP, we study a domain generalization problem where model needs to be generalized to the molecules from different domain splits.
Aligned with the knowledge of biochemistry, we define the following three domains: assay, scaffold and molecular size.
As an illustration, we treat the LBAP problem as a  binary classification problem, where the input $\x$ is the graph data of a small molecule, label $y$ is the ground truth (active or inactive) of binary affinity classification, and 
the $d$ represents domain identifier for one specific domain splits.

\paragraph{Data Info.}
As mentioned before, we preprocess the \chembl dataset and generate in total 36 exemplar datasets with varying noise levels, measurements types, and domain definitions.
Each small molecule in each dataset is represented as a graph, where the nodes are atoms and edges are chemical bonds.
Following the pre-processing strategy in \citep{xiong2019pushing}, we preprocess the molecules via the RDKit package \citep{landrum2013rdkit}. Input node features are 39-dimensional vectors including atomic symbol, hybridization, hydrogens and so on. Input edge features are 10-dimensional vectors including bond type, conjugation, ring and bond stereo chemistry.
Following the detailed splitting strategy in \cref{fig:intro:split} of  \cref{subsec_domain_split_id}, we split the dataset via three types of domain annotations:

\begin{figure}
    \centering
    \scalebarimg{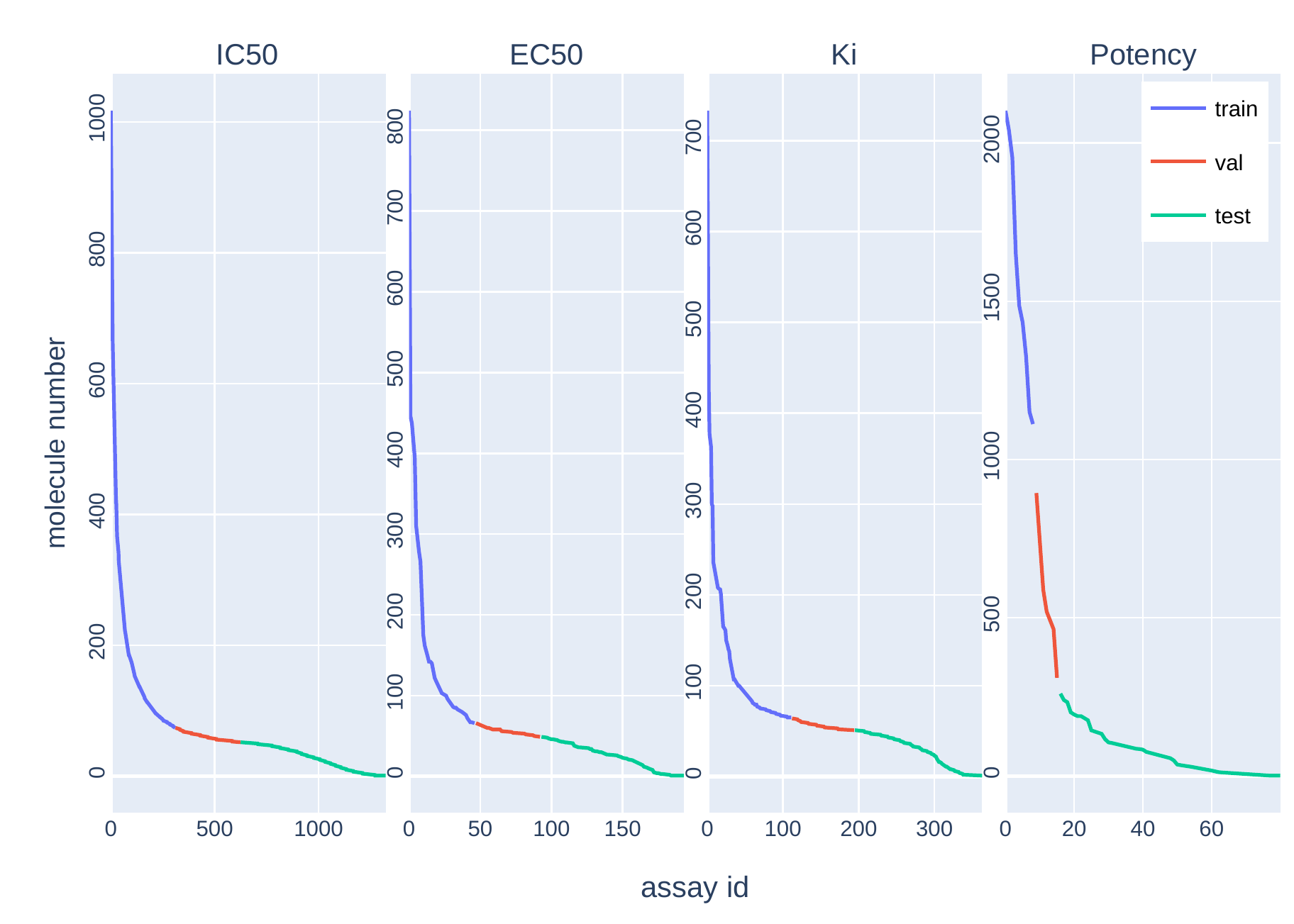}{0.45}{(a)}
    \scalebarimg{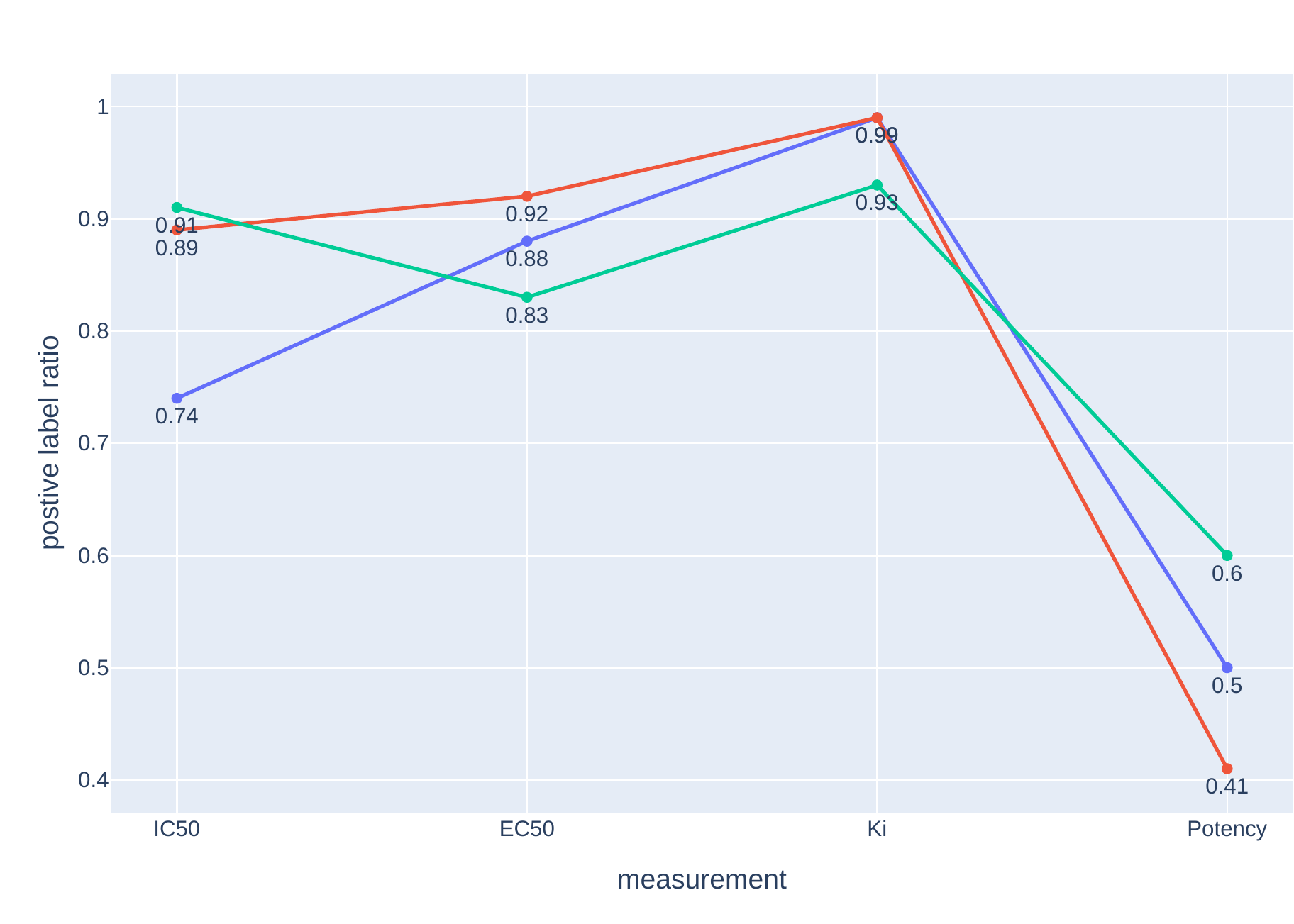}{0.45}{(b)}
    \caption{Analysis of the assay domain in the realized \drugood dataset. (a) shows the distribution of molecule numbers in the assay environment for different measurement types (left to right: IC50, EC50, Ki and Potency)  and (b) shows the  ratio of positive samples of four measurement types in the train/validation/test splits. Note that  here the train, val, test refer to the Intermediate Train, OOD val and OOD test datasets  in \cref{fig:intro:split}.}
    \label{fig:exp:lbap:assay_data_statistic}
\end{figure}

\begin{itemize}
    \item[$\blacktriangleright$] \textbf{Assay:} Samples in the same assay are divided into the same domain. Due to the big differences of experimental environments and protein targets in various assays, the bioactivity values measured by those assays will have a large shift. 
    In this setting, we split the dataset along assays. This split provides a realistic estimate of model performance in prospective experimental settings by separating different molecules into different experimental environments. We assign the assays that contain large number of samples to the training set, and the assays with small number of samples to the test set. After such an assignment, the domain shifts between the training set and the OOD validation/testing set become sufficiently large. The proportion of samples is kept at around 6:2:2 in training, validation, test set. 
    
    Here takes the \drugood-lbap-core-ic50-assay dataset as an illustration here.
    \begin{itemize}
    \item[$\bullet$] \textbf{Train:} 
    Contain total 34,179 molecules from the largest 311 assays with an average of 110 molecules per assay.
    
    \item[$\bullet$] \textbf{Validation (ID):}
    Contain total 11314 molecules from the same 311 assays as in the training sets.
    
    \item[$\bullet$] \textbf{Test (ID):}
    Contain total 11,683 molecules from the same 311 assay environments as in the training sets.    
    
    \item[$\bullet$] \textbf{Validation (OOD):} 
    Contain total 19,028 molecules from the next largest 314 assays with an average of 60.6 molecules per assay.
    
    \item[$\bullet$] \textbf{Test (OOD):} 
    Contain total 19,302 molecules from the smallest 314 assays with an average of 27.6 molecules per assay.
    
    \end{itemize}
    
    \cref{fig:exp:lbap:assay_data_statistic} illustrates the analysis of the assay domain in the realized DrugOOD dataset. As shown in \cref{fig:exp:lbap:assay_data_statistic} (a), the statistics of the assays in terms of the numbers of molecules in each assay, which implies that the scale of the experiments is highly skewed, with the test set containing the assay with the fewest molecules.
    However, the difference in assay environments does not significantly change the statistics of the learning target in each split. In \cref{fig:exp:lbap:assay_data_statistic}, the label statistics still remain very similar in the training/validation/test splits, indicating that the main distribution variation comes from differences in the detection environment.

    \begin{figure}[t]
    \centering
    \scalebarimg{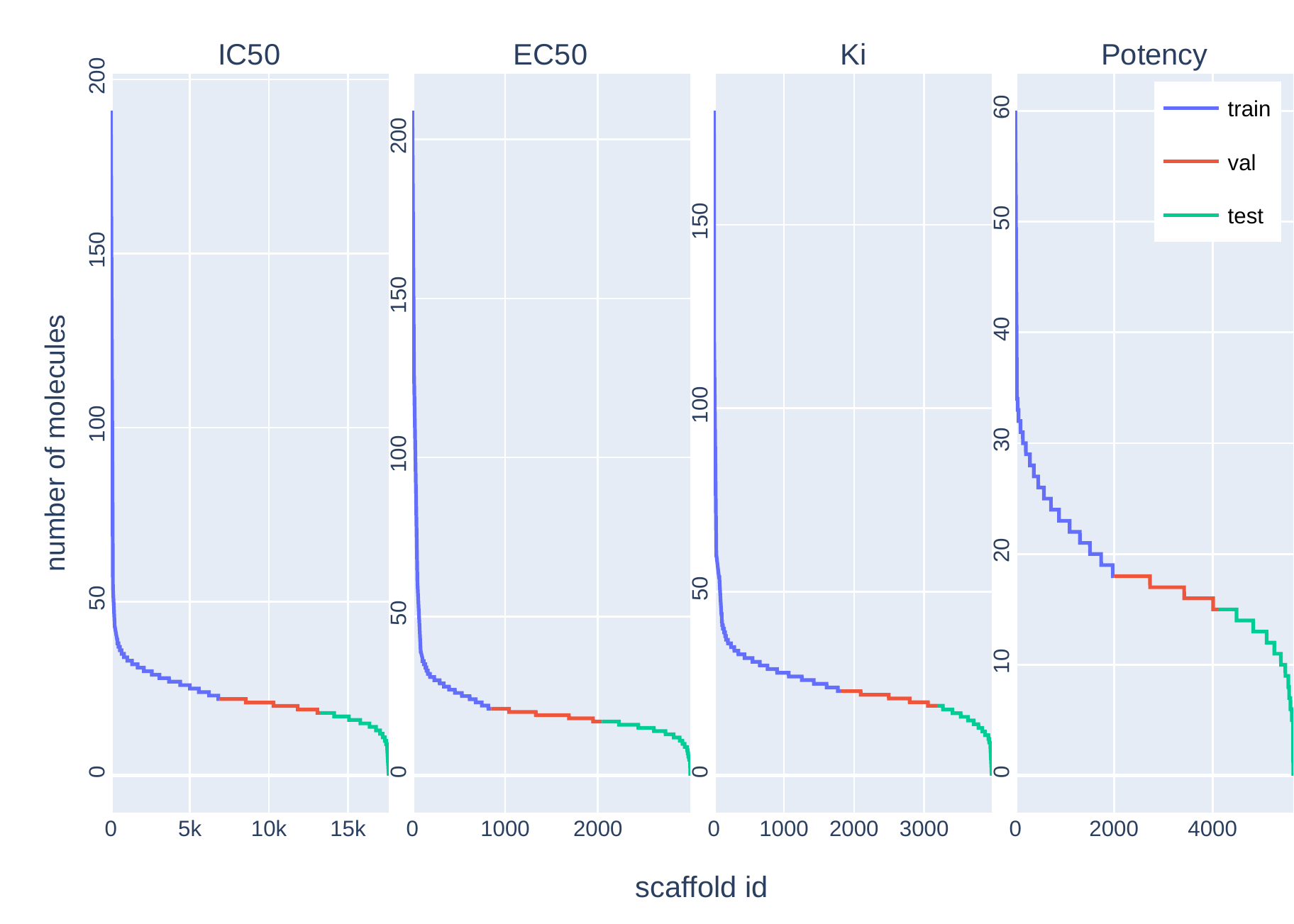}{0.45}{(a)}
    \scalebarimg{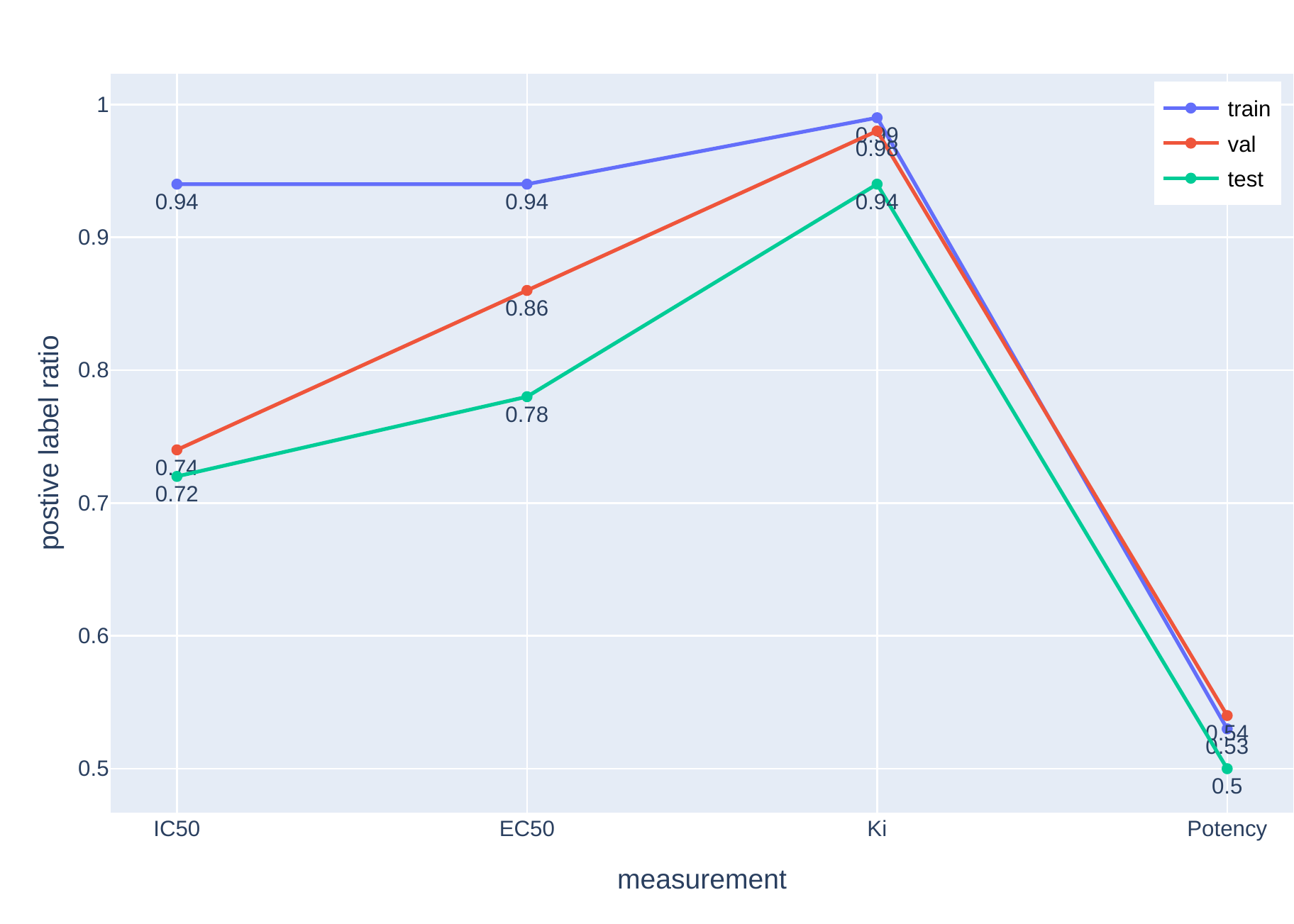}{0.45}{(b)}
    \caption{Analysis of the scaffold domain in the \drugood dataset. (a) shows the distribution of number of molecules wrt the scaffold id and (b) shows the positive ratio of molecules of four measurement types in the train/validation/test splits. Note that  here the train, val, test refer to the Intermediate Train, OOD val and OOD test datasets  in \cref{fig:intro:split}.}
    \label{fig:exp:lbap:scaffold_data_statistic}
    \end{figure}
    
    \item[$\blacktriangleright$] \textbf{Scaffold:}
    Scaffold split has been widely used in previous benchmarks \citep{WILDS, 2020arXiv200500687H}, which splits datasets based on scaffold structure. 
    Similarily, we assign the largest scaffolds to training set and smallest scaffolds to the test set to ensure its maximal diversity in scaffold structure.
    Take the  \drugood-lbap-core-ic50-scaffold dataset as examples, the detailed information of splits are:
    \begin{itemize}
    \item[$\bullet$] \textbf{Train:} 
    Contain in total 21,519 molecules from the largest 6,881 scaffolds with an average of 3.12 molecules per scaffold.
    
    \item[$\bullet$] \textbf{Validation (ID):}
    Contain in total 4,920 molecules from the 1,912 scaffolds with an average of 2.57 molecules per scaffold.
    
    \item[$\bullet$] \textbf{Test (ID):}
    Have 24,112 scaffolds, with in total 30,708 molecules.
    
    \item[$\bullet$] \textbf{Validation (OOD):} 
    The next largest 6,345 scaffolds, with in total 11,683 molecules, and an average of 1.84 molecules per scaffold.

    \item[$\bullet$] \textbf{Test (OOD):} 
    The smallest 4,350 scaffolds, with in total 19,048 molecules, and an average of 4.42 molecules per scaffold.
    \end{itemize}
    
    In the same way, we plot the statics of the scaffolds in terms of the size of each scaffold.
    As shown in \cref{fig:exp:lbap:scaffold_data_statistic}, we again observe that the distribution is highly skewed and the test partition contains the smallest scaffolds.
    
    \item[$\blacktriangleright$] \textbf{Size:}
    We put samples with the same atomic number into a domain, and separate molecules with different atomic sizes into different subsets for simulating the realistic distribution shift.
    We organize the molecules with the largest atomic sizes into the training set and those with smaller atomic sizes are assigned to the test set to ensure sufficient variability of atomic sizes between the training and test data.
   Taking the \drugood-lbap-core-ic50-size dataset dataset as an example, the details of the split are as follows:
   
    \begin{itemize}
    \item[$\bullet$] \textbf{Train:} 
    The largest 190 size groups, with overall 36,597 molecules, and an average of 192.61 molecules per group.
    
    \item[$\bullet$] \textbf{Validation (ID):}
    The 140 size groups with in total 12,153 molecules, and an average of 86.80 molecules per group.
    
    \item[$\bullet$] \textbf{Test (ID):}
    The other 229 groups in the intermediate training data, with total 12411 molecules.
    
    \item[$\bullet$] \textbf{Validation (OOD):} 
    The next largest 4 size groups in addition to id data, with in total 17,660 molecules, and an average of 4,415 molecules per group.

    \item[$\bullet$] \textbf{Test (OOD):} 
    The smallest 18 size groups, with in total 19,048 molecules, and an average of 1,058.22 molecules per group.
    \end{itemize}
    
\end{itemize}

\paragraph{Evaluation.}
We evaluate models' performance by the area under the receiver operating characteristic  (AUROC), which indicates the ability of a classifier to distinguish between classes (e.g., inactive or active). The higher the AUC, the better the performance of the model at distinguishing between the positive and negative classes. Meanwhile, we also provide the results of accuracy metric.

\setlength{\tabcolsep}{5.pt}
\renewcommand{\arraystretch}{1}
\begin{table}[!t]\centering
\caption{Results of the ERM baseline  on datasets: \drugood-lbap-core-ic50-assay (first row), \drugood-lbap-core-ic50-scaffold (second row) and \drugood-lbap-core-ic50-size (third row). In-distribution (ID) results correspond to the train-to-train setting. Parentheses show standard deviation across 3 replicates.}
\scriptsize
\begin{tabular}{cccccccccc} \hline
\multirow{2}{*}{Domain} &\multicolumn{2}{c}{Val (ID)} &\multicolumn{2}{c}{Val (OOD)} &\multicolumn{2}{c}{Test (ID)} &\multicolumn{2}{c}{Test (OOD)} \\
&ACC &AUC &ACC &AUC &ACC &AUC &ACC &AUC \\ \hline \hline
Assay &89.05(0.35) &89.91(1.78) &88.79(2.23) &70.32(0.8) &89.34(0.38) &89.62(2.04) &82.14(0.86) &71.98(0.29) \\
Scaffold &97.04(0.13) &94.84(0.6) &85.11(0.4) &78.96(0.67) &90.51(0.14) &87.15(0.48) &76.33(0.64) &69.54(0.52) \\
Size &93.99(0.04) &92.91(0.16) &84.22(0.22) &81.5(0.19) &93.75(0.06) &92.35(0.15) &71.46(0.67) &67.48(0.47) \\
\hline
\end{tabular}
\label{tab:exp:lbap:erm_drop:b}
\end{table}

\subsubsection{Baseline Results}

\paragraph{Model.}
We train the GIN model as the baseline model on each dataset from scratch with a learning rate at 1e-4, a batch size at 256 samples and without L2-regularization.
Each small molecules are pre-processed by the RDKit package to generate 39-dimensional node features and 9-dimensional edge features as input.
To avoid performance degradation caused by inappropriate hyper-parameters, following the strategy in WILDS \citep{WILDS}, we did a grid search strategy over learning rates $\{0.00003, 0.0001, 0.0005, 0.001, 0.01\}$, batch size $\{64, 128, 256, 512, 1024\}$.
We report averaged results aggregated over 3 random seeds.

\paragraph{ERM results and performance drops.}
As shown in \cref{tab:exp:lbap:erm_drop:a} and \cref{tab:exp:lbap:erm_drop:b}, model performance dropped significantly going from the train-to-train in-distribution (ID) setting to the official out-of-distribution (OOD) setting.
For the assay domain, ERM achieves an average AUC score of 89.91\% on the ID validation set and 89.62\% on the ID test set, but only 70.32\% on the OOD val set, 71.98\% on the OOD test set.
Similarly, for the scaffold and size domain, ERM obtains 87.15\%, 92.35\% AUC score on ID test set but 69.54\%, 67.48\% on OOD test set, respectively.
The test performance of ERM drops by 17.64\%, 17.60\%, 24.97\% points AUC score when the assay, scaffold, size split is used, respectively, suggesting that these splits are indeed harder than conventional random split, and can be used  to estimate the realistic ID-OOD gap in the task of ligand based affinity prediction.
More experimental results on different \drugood lbap datasets are shown in \cref{tab:exp:lbap:erm_drop:c} of the Appendix.

\paragraph{Results of other baselines.}
\cref{tab:exp:lbap:other_baseline} shows the performance of other conventional generalization algorithms. For a fair comparison, all algorithms adopt the same backbone network.
Besides, we also make additional grid searches on algorithms' specific hyper-parameters separately, IRM's penalty weight in $\{1, 10, 100, 1000\}$ and penalty anneal iteration in $\{100, 500, 1000\}$.
DeepCoral's penalty weight in $\{0.1, 1, 10\}$, GroupDro's step size in $\{0.001, 0.01, 0.1\}$,  DANN's inverse factor between $\{0.0001, 1\}$ and Mixup's probability and interpolate strength between $\{0.0001, 1\}$.

As shown in \cref{tab:exp:lbap:other_baseline}, ERM almost performs better than DeepCoral, IRM, and Group DRO, all of which use assay, scaffolds, size as the domains, indicating these existing methods can not solve the \drugood problem.
Moreover, similar to the findings in the WILDS benchmark, current existing methods make model hard to fit the training data, For instance, under the scaffold domain split, DeepCoral, IRM achieves 81.6\%, 77.66\% AUC score in the In-Distribution validation set, respectively, while the AUC score pf ERM baseline is 94.84\%.
Also, these methods are primarily designed for the case when each group contains a decent number of examples, which is not the common case for the drug development scenario.
Finally, the  SOTA OOD algorithm does not work well in the \drugood setting, suggesting that robust methods need to be developed to solve the OOD problem for graph data.

\setlength{\tabcolsep}{2pt}
\renewcommand{\arraystretch}{1}
\begin{table}[!t]\centering
\scriptsize
\caption{Baseline results on three datasets for the six OOD algorithm.  \drugood-lbap-core-ic50-assay (first row), \drugood-lbap-core-ic50-scaffold (second row) and \drugood-lbap-core-ic50-size (third row). In-distribution (ID) results correspond to the train-to-train setting. Parentheses show standard deviation across 3 replicates.}
\begin{tabular}{ccccccccccc} \hline
\multirow{2}{*}{Domain} &\multirow{2}{*}{Algos} &\multicolumn{2}{c}{Val (ID)} &\multicolumn{2}{c}{Val (OOD)} &\multicolumn{2}{c}{Test (ID)} &\multicolumn{2}{c}{Test (OOD)} \\
& &ACC &AUC &ACC &AUC &ACC &AUC &ACC &AUC \\ \hline \hline
\multirow{6}{*}{Assay} &ERM&89.05 (0.35) &89.91 (1.78) &88.79 (2.23) &70.32 (0.80) &89.34 (0.38) &89.62 (2.04 &82.14 (0.86) &71.98 (0.29)\\
&IRM&88.14 (0.17) &82.82 (0.87) &90.67 (0.07) &68.23 (0.31) &88.39 (0.25) &83.10 (0.46 &82.41 (0.20) &69.22 (0.51)\\
&DeepCoral&88.59 (0.10) &88.10 (1.42) &91.09 (0.15) &70.26 (1.04) &88.88 (0.15) &88.23 (1.42 &83.04 (0.08) &71.76 (0.60)\\
&DANN&88.47 (0.20) &83.39 (1.15) &91.32 (0.38) &68.30 (0.22) &88.72 (0.16) &83.20 (1.28 &83.22 (0.10) &70.08 (0.65)\\
&Mixup&88.80 (0.52) &89.01 (2.06) &88.76 (1.92) &69.14 (0.56) &89.01 (0.43) &88.95 (2.17 &81.65 (1.06) &71.34 (0.41)\\
&GroupDro&88.80 (0.15) &89.42 (0.43) &89.81 (0.41) &70.34 (0.91) &88.96 (0.15) &89.24 (0.82 &82.62 (0.23) &71.54 (0.46)\\ \hline
\multirow{6}{*}{Scaffold}
&ERM&97.04 (0.13) &94.84 (0.60) &85.11 (0.40) &78.96 (0.67) &90.51 (0.14) &87.15 (0.48 &76.33 (0.64) &69.54 (0.52)\\
&IRM&92.51 (4.34) &77.66 (0.79) &81.52 (4.25) &72.77 (0.66) &86.99 (3.93) &77.22 (0.32 &72.96 (4.32) &64.94 (0.30)\\
&DeepCoral&95.76 (0.29) &81.60 (1.45) &85.37 (0.47) &77.09 (0.26) &89.96 (0.08) &81.13 (0.49 &76.90 (0.37) &68.54 (0.01)\\
&DANN&95.89 (0.09) &77.09 (0.64) &85.13 (0.81) &75.04 (0.65) &89.86 (0.10) &77.30 (0.65 &77.11 (0.66) &66.37 (0.20)\\
&Mixup&97.19 (0.08) &95.51 (0.44) &85.55 (0.08) &79.42 (0.62) &90.74 (0.11) &87.35 (0.33 &77.18 (0.19) &69.29 (0.24)\\
&GroupDro&96.02 (0.12) &78.67 (2.75) &85.01 (0.64) &74.57 (0.60) &89.87 (0.18) &78.32 (1.09 &76.18 (0.86) &66.67 (0.67)\\
\hline
\multirow{6}{*}{Size} 
&ERM&93.99 (0.04) &92.91 (0.16) &84.22 (0.22) &81.50 (0.19) &93.75 (0.06) &92.35 (0.15 &71.46 (0.67) &67.48 (0.47)\\
&IRM&86.77 (4.05) &66.41 (1.79) &76.49 (4.36) &60.59 (0.29) &87.08 (3.72) &69.80 (1.74 &66.39 (3.92) &57.00 (0.39)\\
&DeepCoral&92.40 (0.06) &70.70 (1.68) &83.34 (0.27) &61.90 (0.39) &92.28 (0.20) &73.08 (0.98 &72.36 (0.32) &57.31 (0.44)\\
&DANN&91.31 (2.12) &80.13 (3.59) &81.99 (2.46) &73.73 (0.49) &91.12 (2.12) &78.53 (3.71 &70.08 (3.50) &63.45 (0.18)\\
&Mixup&94.23 (0.21) &92.89 (0.30) &84.64 (0.26) &81.79 (0.20) &93.88 (0.17) &92.48 (0.55 &72.73 (0.68) &67.73 (0.27)\\
&GroupDro&92.73 (0.17) &78.46 (0.91) &83.54 (0.52) &67.68 (0.42) &92.67 (0.27) &80.02 (0.44 &72.64 (0.33) &60.90 (1.23)\\
\bottomrule
\end{tabular}
\label{tab:exp:lbap:other_baseline}
\end{table}

\begin{figure}[]
\centering
\begin{minipage}[t]{0.32\columnwidth}
\centering
\includegraphics[width=\columnwidth]{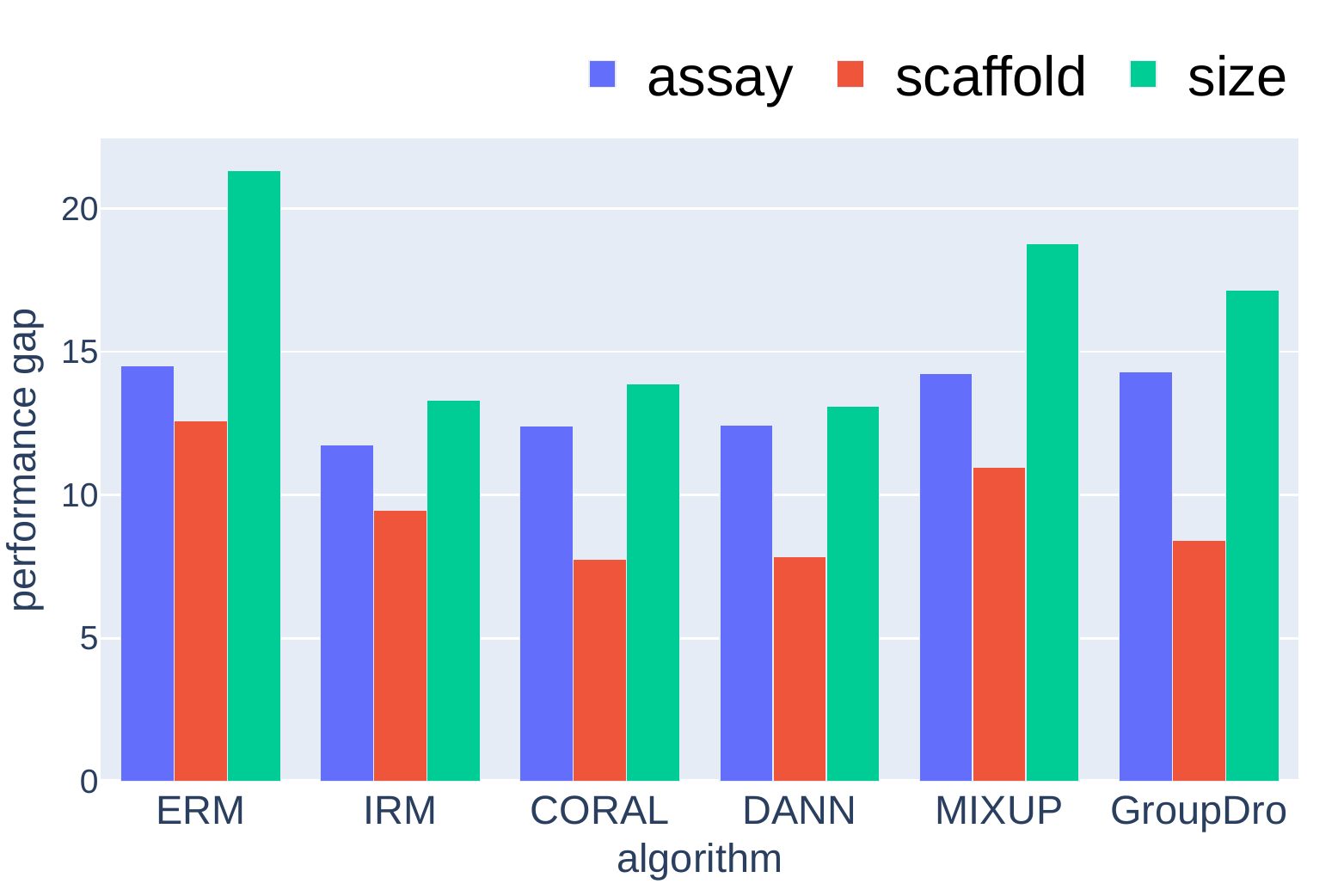}
\end{minipage}
\begin{minipage}[t]{0.32\columnwidth}
\centering
\includegraphics[width=\columnwidth]{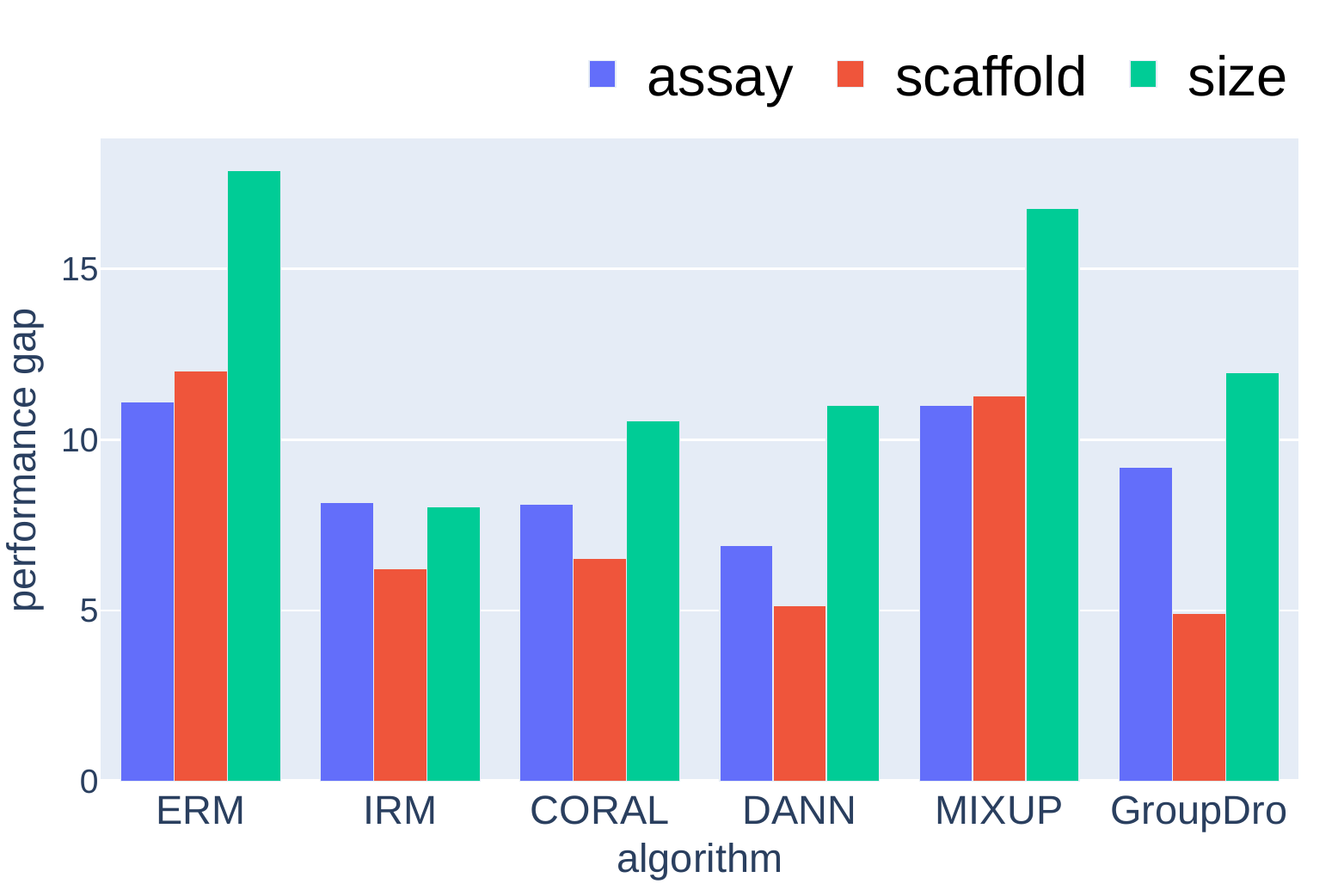}
\end{minipage}
\begin{minipage}[t]{0.32\columnwidth}
\centering
\includegraphics[width=\columnwidth]{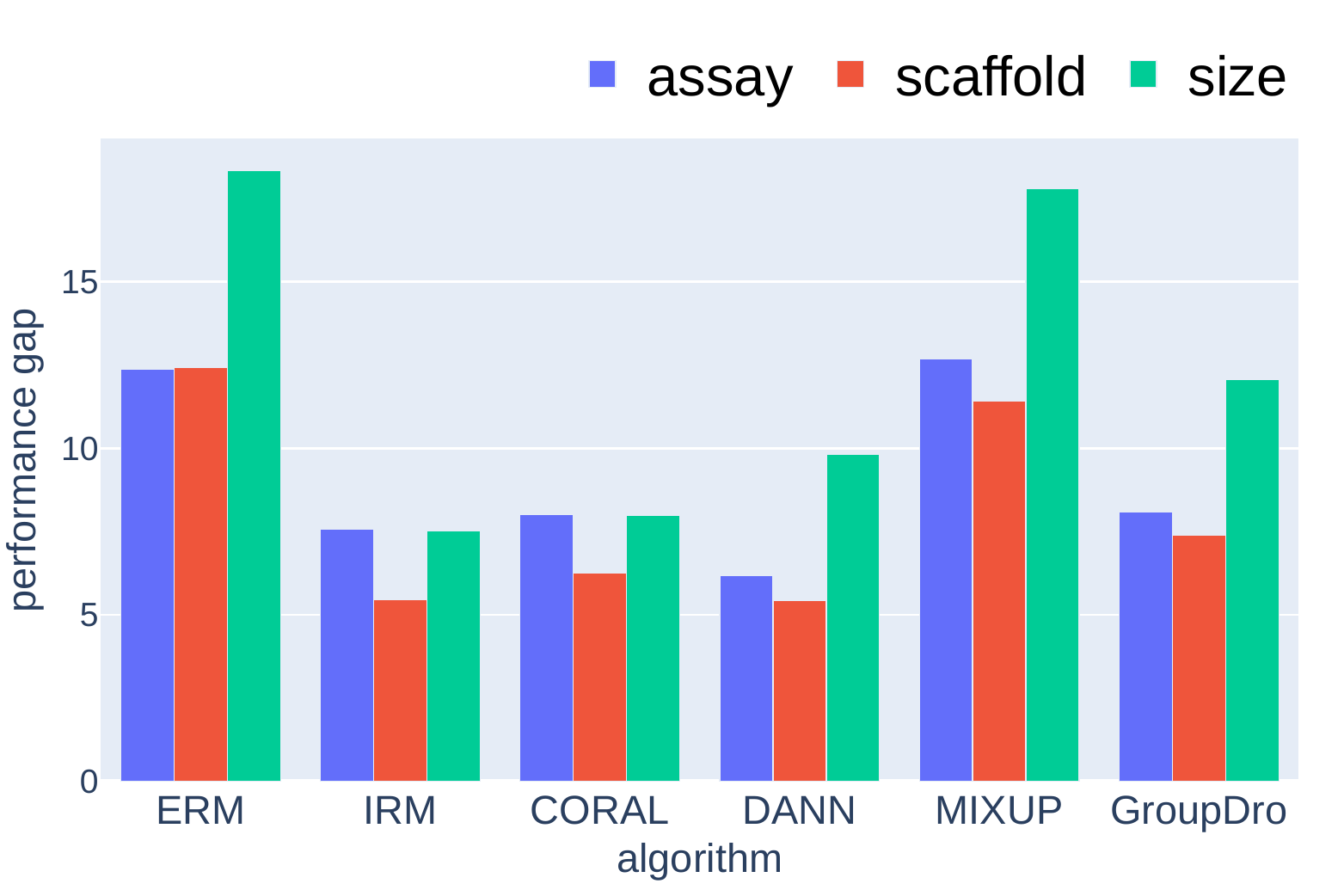}
\end{minipage}
\caption{Performance gap on AUC of different domain splits with varying levels of noise and OOD algorithms on the \drugood-lbap datasets. The gap values are calculated on the OOD test set and averaged over four measurement types. From left to right: core noise level, refined noise level, general noise level. Color indicates domain annotations for LBAP tasks.}
\label{fig:exp:lbap:domain_split}
\end{figure}

\subsubsection{Performance Drops of Different Domains}
For the LBAP task, three types of domains are defined, and we study the performance of the model under these different domain splits.
\cref{fig:exp:lbap:domain_split} shows the performance degradation for different domain partitions, the gap values are computed on the test set and averaged over all measurement types. 
From the figure, we can conclude the following.
1) Among several divisions, the size domain brings the largest performance degradation, which is consistent with daily experimental findings, where molecules of different sizes often have very different properties.
2) As the noise level of the data set increases, the performance degradation of different domains is somewhat mitigated, and the increased data to some extent increase the generalization ability of the model. In addition, the degradation of the model performance is not further mitigated as the data continues to increase, indicating that the increase in the amount of data brings limited improvement and that a truly effective method needs to be developed to address the \drugood problem.

\subsubsection{Performance Drops of Different Noise Levels}
We also investigated the ID and OOD performance at different noise levels.
As shown in \cref{tab:exp:lbap:noise:drop}, we summarize different algorithms' ID and OD performance under three noise levels.
From the table, one can observe that: 
1) In the presence of more noise, the introduced noise produces pollution, which progressively affects the network's performance.
2) The increasing noise level brings more data, which provides more information about the dataset.
From the core to refined level, we can see a narrowing of the gap between ID and OOD, however, the improvement from the refined level to general level reaches a bottleneck without significant improvement.
3) By combining the above two points, we can see that the introduction of large amounts of data with noise affects the learning of the model to a certain degree and affects its performance, but these noisy data, in turn, provide additional information so that the model is able to improve a small amount of generalization ability.

\setlength{\tabcolsep}{2pt}
\renewcommand{\arraystretch}{1}
\begin{table}[!t]\centering
\caption{Baseline results on datasets: \drugood-lbap-core-ic50-assay (first row), \drugood-lbap-refined-ic50-assay (second row) and \drugood-lbap-general-ic50-size (third row) dataset. In-distribution (ID) results correspond to the train-to-train setting. Parentheses show standard deviation across 3 replicates.}
\scriptsize
\begin{tabular}{lcccccccccc}\toprule
\multirow{2}{*}{Noise} &\multirow{2}{*}{Algos} &\multicolumn{2}{c}{Val (ID)} &\multicolumn{2}{c}{Val (OOD)} &\multicolumn{2}{c}{Test (ID)} &\multicolumn{2}{c}{Test (OOD)} \\
& &ACC &AUC &ACC &AUC &ACC &AUC &ACC &AUC \\ \hline \hline
\multirow{6}{*}{Core} 
&ERM&89.05 (0.35) &89.91 (1.78) &88.79 (2.23) &70.32 (0.80) &89.34 (0.38) &89.62 (2.04 &82.14 (0.86) &71.98 (0.29)\\
&IRM&88.14 (0.17) &82.82 (0.87) &90.67 (0.07) &68.23 (0.31) &88.39 (0.25) &83.10 (0.46 &82.41 (0.20) &69.22 (0.51)\\
&DeepCoral&88.59 (0.10) &88.10 (1.42) &91.09 (0.15) &70.26 (1.04) &88.88 (0.15) &88.23 (1.42 &83.04 (0.08) &71.76 (0.60)\\
&DANN&88.47 (0.20) &83.39 (1.15) &91.32 (0.38) &68.30 (0.22) &88.72 (0.16) &83.20 (1.28 &83.22 (0.10) &70.08 (0.65)\\
&Mixup&88.80 (0.52) &89.01 (2.06) &88.76 (1.92) &69.14 (0.56) &89.01 (0.43) &88.95 (2.17 &81.65 (1.06) &71.34 (0.41)\\
&GroupDro&88.80 (0.15) &89.42 (0.43) &89.81 (0.41) &70.34 (0.91) &88.96 (0.15) &89.24 (0.82 &82.62 (0.23) &71.54 (0.46)\\ \hline
\multirow{6}{*}{Refined} 
&ERM&91.66 (0.11) &89.06 (0.63) &91.61 (0.52) &75.57 (0.16) &91.68 (0.08) &89.25 (0.64 &84.85 (0.33) &72.70 (0.00)\\
&IRM&91.36 (0.08) &84.67 (0.91) &92.24 (0.13) &75.39 (0.32) &91.47 (0.08) &84.80 (0.95 &85.17 (0.23) &71.68 (0.53)\\
&DeepCoral&91.40 (0.05) &84.63 (0.85) &92.21 (0.23) &75.72 (0.47) &91.43 (0.04) &84.55 (0.84 &85.23 (0.13) &71.45 (0.09)\\
&DANN&91.21 (0.12) &78.34 (0.20) &92.27 (0.31) &72.12 (0.39) &91.33 (0.16) &78.40 (0.24 &85.09 (0.34) &67.20 (0.68)\\
&Mixup&91.47 (0.03) &87.30 (0.63) &92.25 (0.02) &75.67 (0.22) &91.57 (0.03) &87.36 (0.61 &85.16 (0.11) &72.13 (0.21)\\
&GroupDro&91.33 (0.10) &84.39 (1.72) &92.15 (0.28) &75.28 (0.23) &91.46 (0.15) &84.55 (2.20 &85.20 (0.32) &71.40 (0.52)\\
\hline
\multirow{6}{*}{General}
&ERM&88.91 (0.14) &85.17 (1.19) &82.48 (0.07) &74.04 (0.01) &89.21 (0.13) &85.19 (1.15 &71.85 (0.08) &69.88 (0.13)\\
&IRM&88.37 (0.05) &80.77 (0.11) &82.37 (0.08) &73.11 (0.10) &88.67 (0.04) &80.75 (0.03 &71.51 (0.07) &69.21 (0.16)\\
&DeepCoral&88.41 (0.10) &81.01 (0.75) &82.43 (0.12) &73.57 (0.25) &88.71 (0.13) &81.00 (0.83 &71.65 (0.08) &69.47 (0.13)\\
&DANN&87.77 (0.40) &76.47 (0.71) &81.87 (0.53) &70.40 (0.57) &88.08 (0.38) &75.82 (0.57 &70.73 (0.54) &65.12 (0.94)\\
&Mixup&89.11 (0.31) &86.00 (1.15) &82.22 (0.28) &74.17 (0.02) &89.47 (0.23) &85.95 (1.19 &71.74 (0.07) &69.87 (0.17)\\
&GroupDro&88.35 (0.17) &81.06 (0.39) &82.39 (0.17) &73.45 (0.33) &88.68 (0.17) &81.31 (0.34 &71.55 (0.06) &69.36 (0.09)\\
\bottomrule
\end{tabular}
\label{tab:exp:lbap:noise:drop}
\end{table}

\subsection{Structure Based Affinity Prediction (SBAP)}

\subsubsection{Setup}
Compared with the LBAP task, the SBAP task considers target protein information. In our benchmark, we represent proteins as protein sequences cause the \chembl database only provide protein sequence information. However,
\drugood can be easily extended by incorporating 3D structure information of targets by refering to 
protein sturcute   depositing database, such as PDB and uniprot \citep{UniProt}.  This will be left as important future work. 

\setlength{\tabcolsep}{2pt}
\renewcommand{\arraystretch}{1}
\begin{table}[]
\caption{Baseline results on datasets: \drugood-sbap-core-ic50-protein (first row), \drugood-sbap-refined-ic50-protein (second row) and \drugood-sbap-general-ic50-protein (third row) dataset. In-distribution (ID) results correspond to the train-to-train setting. Parentheses show standard deviation across three replicates.}
\scriptsize
\centering
\begin{tabular}{cccccccccc} \hline 
\multirow{2}{*}{Noise} &
\multirow{2}{*}{Algos} & \multicolumn{2}{c}{ID\_VAL} & \multicolumn{2}{c}{OOD\_VAL} & \multicolumn{2}{c}{ID\_TEST} & \multicolumn{2}{c}{OOD\_TEST} \\ 
                 &         & ACC          & AUC          & ACC          & AUC          & ACC           & AUC          & ACC           & AUC          \\ \hline \hline
\multirow{4}{*}{Core} 
&ERM&89.28 (0.28) &90.83 (0.48) &84.22 (1.13) &74.12 (0.41) &89.32 (0.19) &90.71 (0.29 &80.37 (1.32) &68.87 (0.53)\\
&IRM&88.94 (0.67) &88.93 (4.03) &82.61 (2.21) &74.10 (0.39) &89.04 (0.60) &88.81 (4.03 &77.92 (4.25) &66.31 (0.45)\\
&DeepCoral&89.03 (0.62) &90.86 (0.95) &82.11 (1.97) &74.12 (0.46) &89.17 (0.76) &90.65 (0.97 &78.29 (2.39) &67.56 (1.01)\\
&DANN&88.23 (0.01) &84.58 (0.34) &84.93 (0.26) &71.47 (0.47) &88.54 (0.10) &84.49 (0.25 &81.93 (0.51) &67.76 (0.41)\\
&Mixup&89.11 (0.15) &90.83 (0.38) &83.52 (2.61) &73.41 (0.73) &89.52 (0.20) &90.94 (0.32 &80.37 (2.76) &67.97 (0.15)\\
&GroupDro&89.05 (0.52) &89.49 (2.92) &83.26 (1.57) &74.10 (0.14) &89.22 (0.35) &89.26 (2.79 &79.63 (2.41) &68.07 (0.98) \\
\hline
\multirow{6}{*}{Refined} 
&ERM&91.99 (0.26) &87.11 (1.50) &85.02 (0.31) &74.41 (0.09) &92.11 (0.20) &86.87 (1.41 &83.17 (0.66) &69.51 (0.30)\\
&IRM&92.40 (0.12) &89.66 (0.63) &84.12 (0.14) &74.47 (0.15) &92.46 (0.13) &89.49 (0.74 &81.73 (0.21) &69.30 (0.48)\\
&DeepCoral&92.00 (0.03) &86.24 (0.15) &85.16 (0.33) &74.56 (0.25) &92.16 (0.09) &85.74 (0.24 &83.41 (0.40) &69.27 (0.52)\\
&DANN&91.61 (0.13) &78.02 (0.54) &85.23 (0.17) &71.29 (0.17) &91.88 (0.15) &77.14 (0.64 &83.42 (0.41) &66.58 (0.29)\\
&Mixup&92.01 (0.29) &86.58 (1.84) &85.39 (0.08) &74.25 (0.16) &92.18 (0.15) &86.02 (1.95 &83.76 (0.13) &69.29 (0.19)\\
&GroupDro&92.06 (0.17) &87.14 (0.47) &85.03 (0.16) &74.30 (0.22) &92.26 (0.09) &86.73 (0.33 &83.22 (0.42) &69.40 (0.40)\\
\hline
\multirow{6}{*}{General} 
&ERM&91.16 (0.18) &85.62 (1.66) &85.61 (0.74) &73.67 (0.08) &90.95 (0.13) &85.34 (1.67 &80.69 (0.95) &68.48 (0.27)\\
&IRM&90.87 (0.19) &82.78 (2.30) &85.81 (0.11) &73.61 (0.18) &90.70 (0.12) &82.68 (2.40 &80.90 (0.34) &68.43 (0.33)\\
&DeepCoral&90.82 (0.10) &82.22 (1.96) &86.00 (0.13) &73.34 (0.06) &90.63 (0.07) &81.98 (1.94 &81.04 (0.60) &68.13 (0.73)\\
&DANN&90.84 (0.03) &80.96 (1.18) &85.90 (0.54) &71.52 (0.24) &90.70 (0.03) &80.20 (1.03 &81.08 (0.82) &66.38 (0.29)\\
&Mixup&90.98 (0.18) &83.05 (1.63) &86.11 (0.20) &73.49 (0.28) &90.81 (0.19) &82.88 (1.70 &81.32 (0.39) &68.45 (0.14)\\
&GroupDro&90.77 (0.09) &82.53 (1.30) &85.69 (0.54) &73.36 (0.40) &90.59 (0.10) &82.43 (1.30 &80.49 (1.07) &68.11 (0.38)\\
\bottomrule
\end{tabular}
\label{tab:sbap_IC50_protein}
\end{table}

\subsubsection{Model}
We use a general SBAP prediction network that extracts molecular and protein features separately, which are then concatenated together and fed into a fully connected layer to predict interaction probabilities.
For the feature extraction of small molecules, we follow the setting of LBAP task and use the same network and hyperparameters.
For protein sequence, we use the pre-train BERT~\citep{devlin2018bert}: 'bert-base-uncase' to 
extract a 768 dimensional protein feature.
Then, the features of molecules and protein are concatenated and fed into a one-layer fully connected layer to predict the interaction probabilities.

\subsubsection{Baseline Results}
The results of different algorithms on sub-datasets with IC50 as measurement type and protein as domain are shown in \cref{tab:sbap_IC50_protein}.
From the table, we can see that:
1) The performance of OOD is degraded relative to that of ID. On the validation set of core noise level, the OOD performance of ERM degrades by 16.71\% in AUC relative to the ID performance, while the performance drop expanded to 21.84\% on test set.
2) OOD algorithms designed for computer vision tasks hardly work in SBAP scenarios. The performance of algorithms designed for OOD scenarios are difficult to match the performance of ERM, which means in order to promote the development of AI aided drug discovery, it is necessary to design OOD algorithms with the consideration of characteristics of drug development scenarios.
Next, we will further analyze the experimental results from different aspects.

\subsubsection{Performance Gap of Different Domains}
\begin{figure}[]
\centering
\begin{minipage}[t]{0.325\columnwidth}
\centering
\includegraphics[width=1\linewidth]{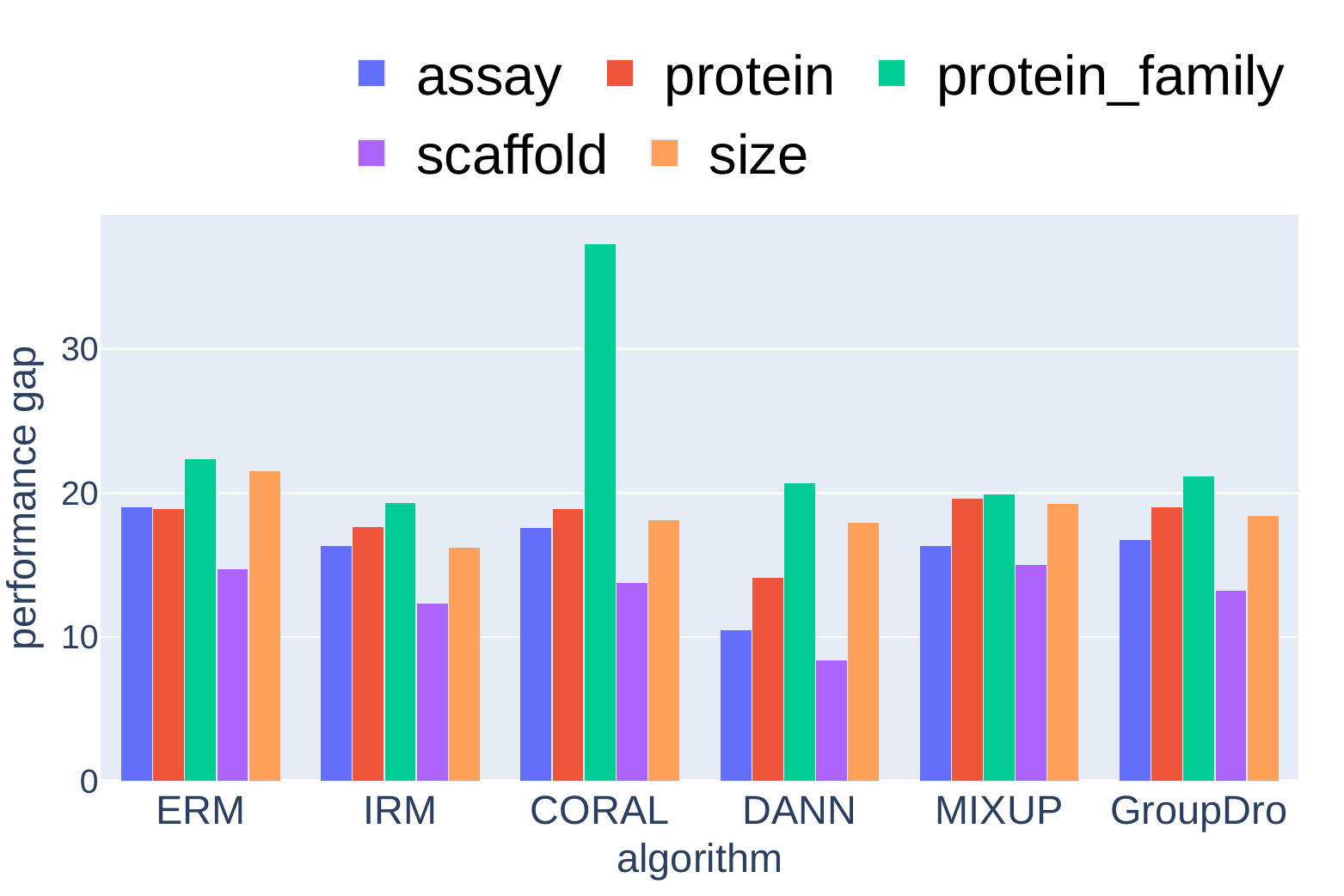}
\end{minipage}
\begin{minipage}[t]{0.325\columnwidth}
\centering
\includegraphics[width=1\linewidth]{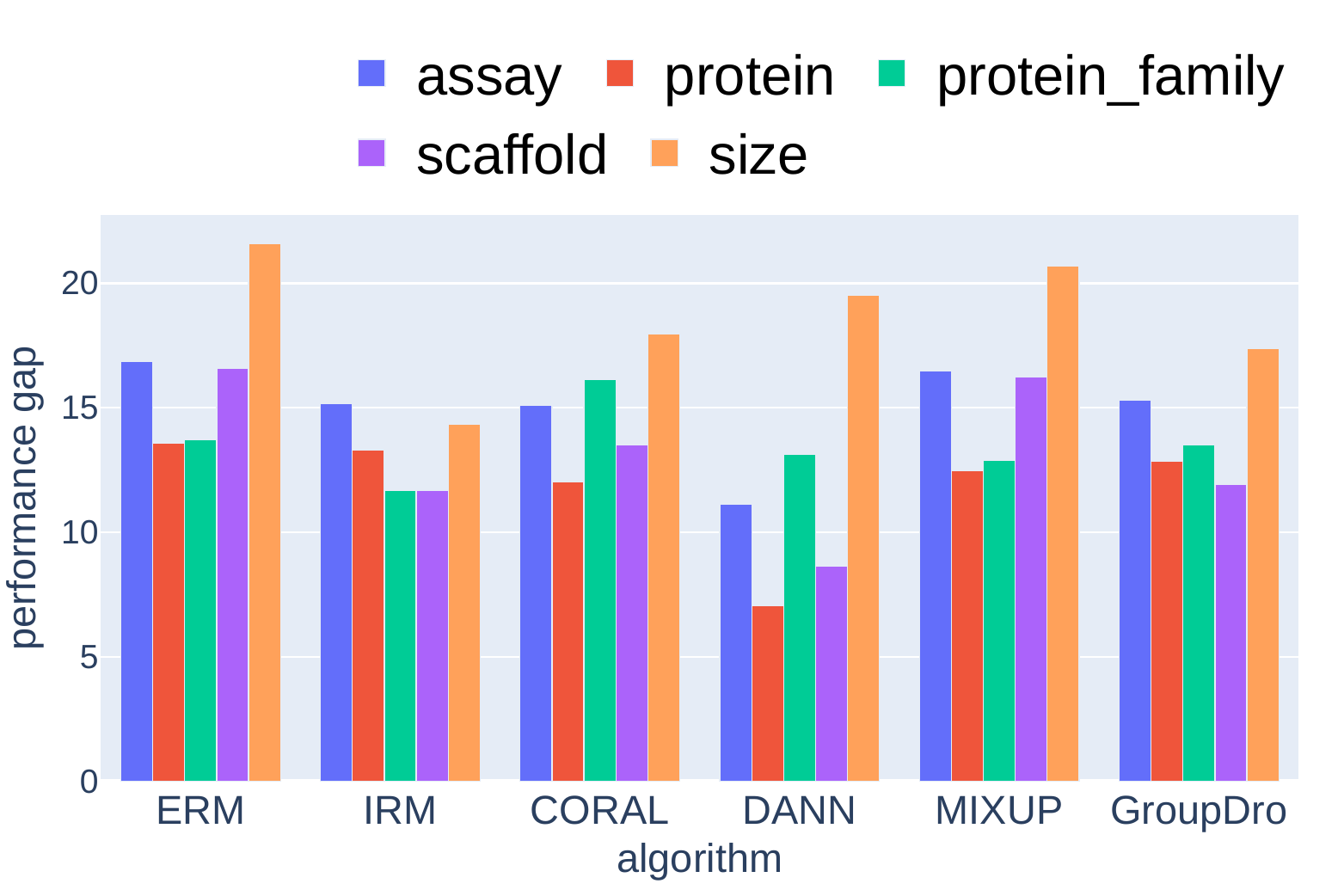}
\end{minipage}
\begin{minipage}[t]{0.325\columnwidth}
\centering
\includegraphics[width=1\linewidth]{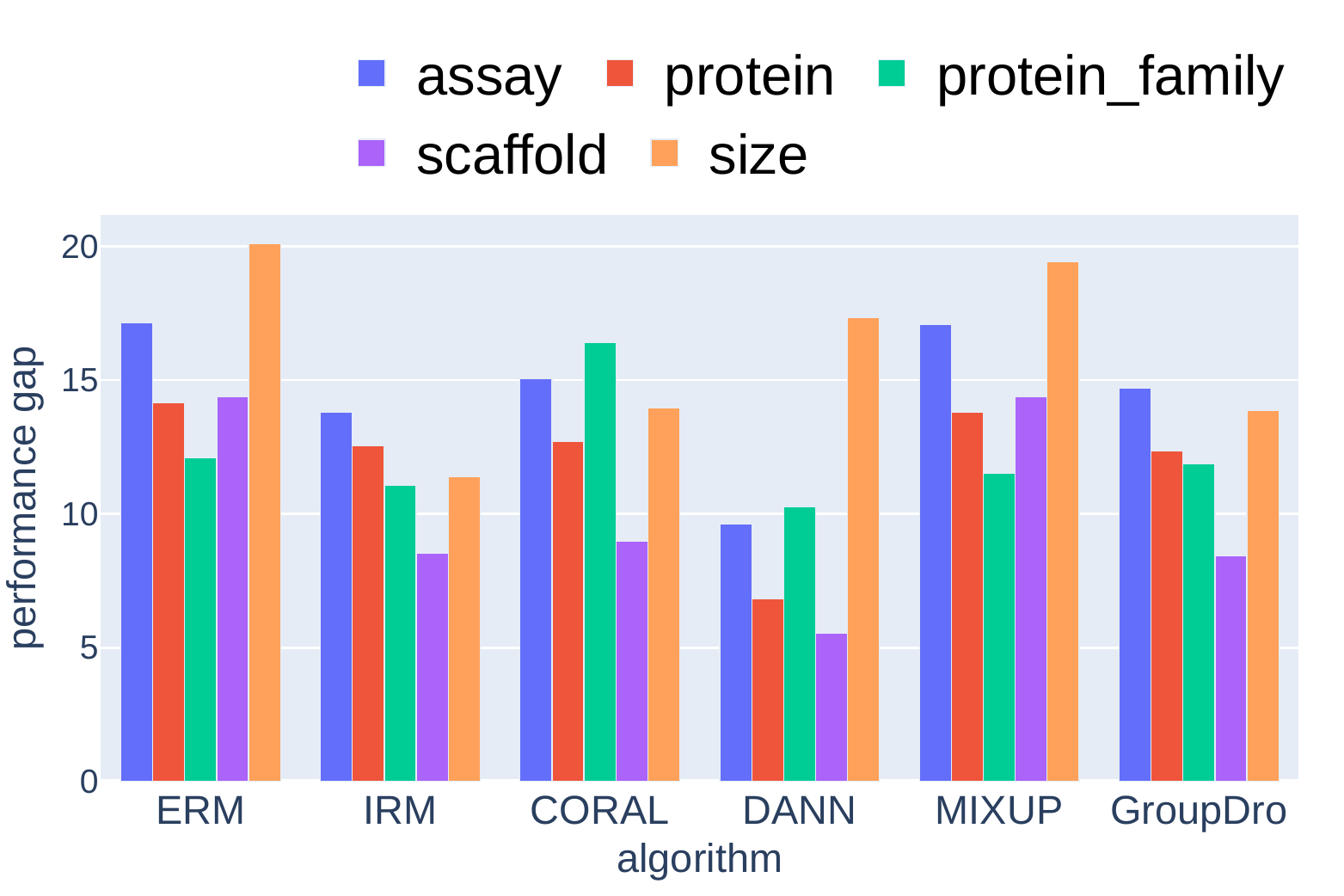}
\end{minipage}
\caption{Performance gap on AUC of different domain splits with different noise levels and OOD algorithms on the \drugood-sbap datasets. The gap values are calculated on the OOD test set and averaged over four measurement types.  Left to right: core noise level, refined noise level, general noise level.  Color indicates domain splits for SBAP tasks.  
}
\label{fig:performance_gap_domains}
\end{figure}

Different domain split methods may result in different distribution shifts, and therefore will also bring different challenges to the algorithms.
Here, we analyze 5 built-in domain splits for the SBAP task.
\cref{fig:performance_gap_domains} shows the performance gap of different domain splits. We conduct analysis under different noise levels and algorithms. The values of gap are calculated on the testing set and averaged over all measurements types.
From ~\cref{fig:performance_gap_domains}, we can see that:
1) The noise level of the dataset has a great impact on the performance gap of ID and OOD. 
On one hand, each noise level has a different domain split method that can bring the biggest performance gap across all algorithms. For example, in core noise level, the domain 'protein family' brings the largest performance gap across all algorithms, while it changes to be 'size' in refined noise level.
On the other hand, as the noise level increases, the performance gap gradually narrows. Combining with ~\cref{tab:sbap_IC50_protein}, we observe that this is caused by the degradation of ID performance.
2) Among all domain split methods, the gap brought by scaffold is relatively small. We speculate that this may be due to the fact that scaffold split induces more domain than other split methods, which makes the training set cover a wider portion of the underlying distribution.

\begin{figure}[!h]
    \centering
    \includegraphics[width=0.85\columnwidth]{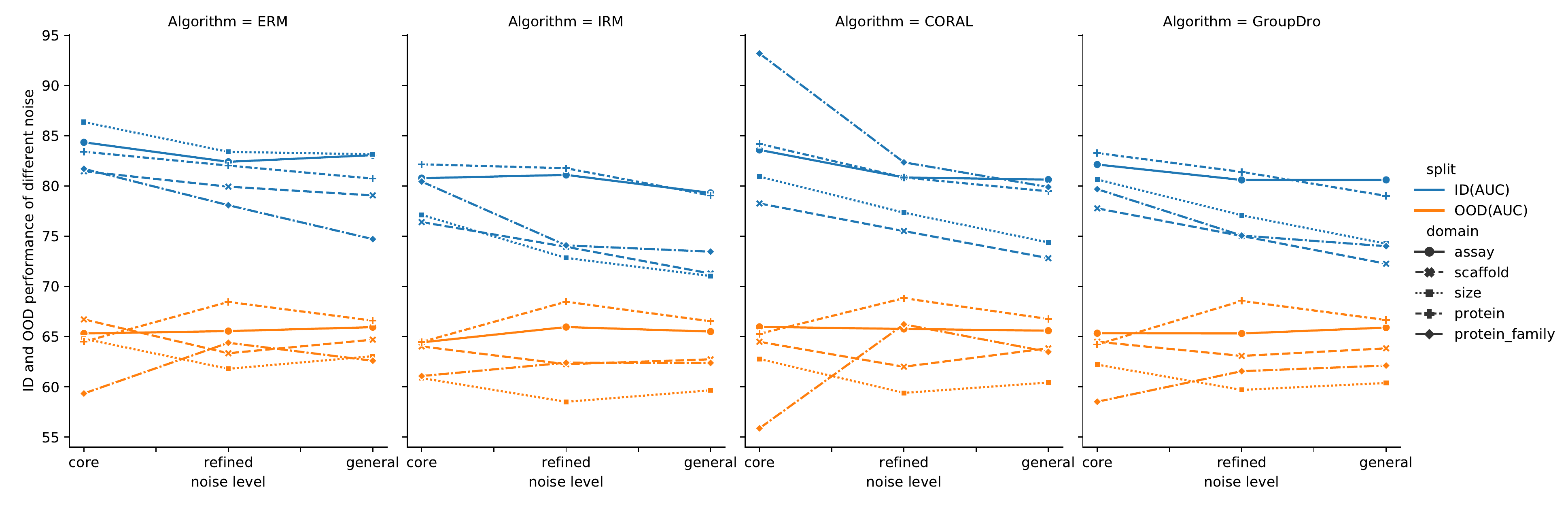}
    \caption{ID and OOD performance with different noise levels. Left to right: ERM, IRM, DeepCoral, GroupDro. Blue color indicates the ID performance while orange color denotes the OOD performance. }
    \label{fig:exp_sbap_ID_OOD_noise}
\end{figure}

\subsubsection{Study for Different Noise Levels}
Here, we study the effect of different noise levels on ID and OOD performance.
~\cref{fig:exp_sbap_ID_OOD_noise} shows the ID and OOD performance of 4 algorithms with different noises. The values are averaged over all measurement types.
It can be seen that the performance of ID generally decreases with the increase of noise level. The introduction of noise hurts the performance of the model despite the larger amount of data, which suggests that we need to pay more attention to the noise of the data source in the realistic scenario. How to design an algorithm that is robust to both noise and distribution shift is a worthy research direction.
As for OOD performance, we can see that different domain split shows different changing trend while the noise level increases.
However, we notice that the OOD performance of different algorithms under a particular domain exhibits similar behavior for ERM, which means that existing OOD algorithms do not incorporate the noise of the real scene, nor are they designed to mitigate its impact.

\begin{figure}[!t]
\centering
\scalebarimg{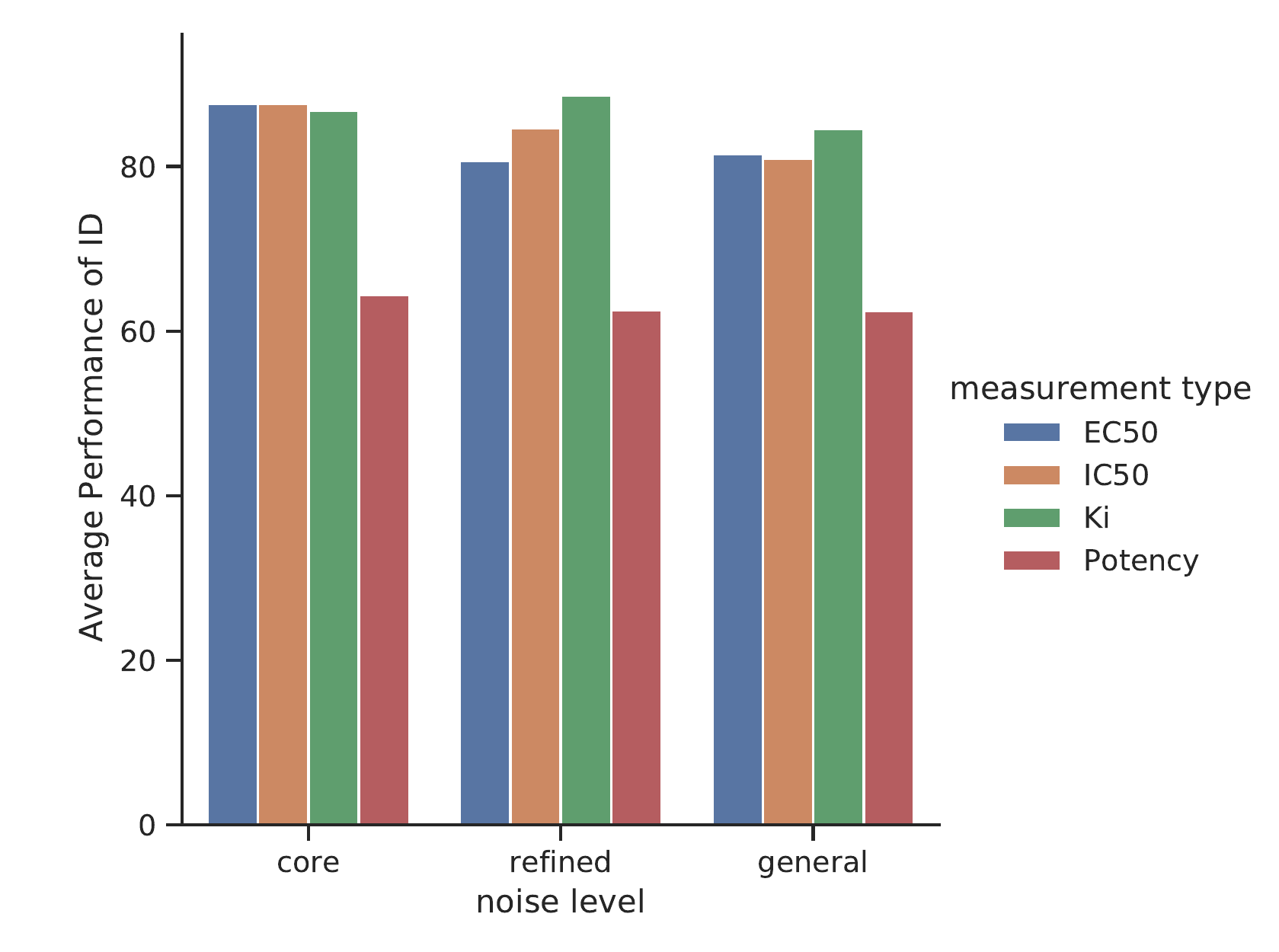}{0.35}{(a)}
\scalebarimg{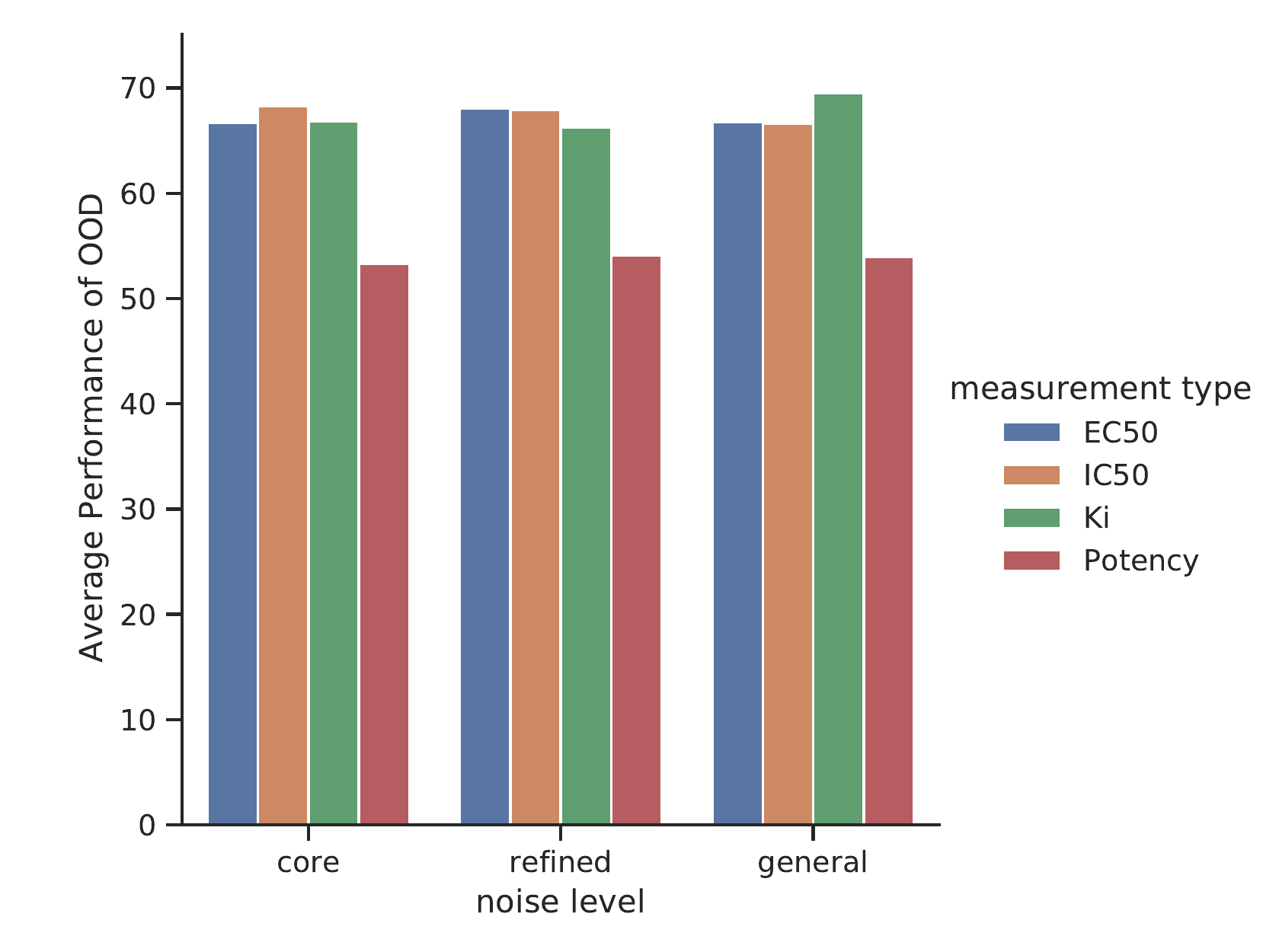}{0.35}{} \\
\scalebarimg{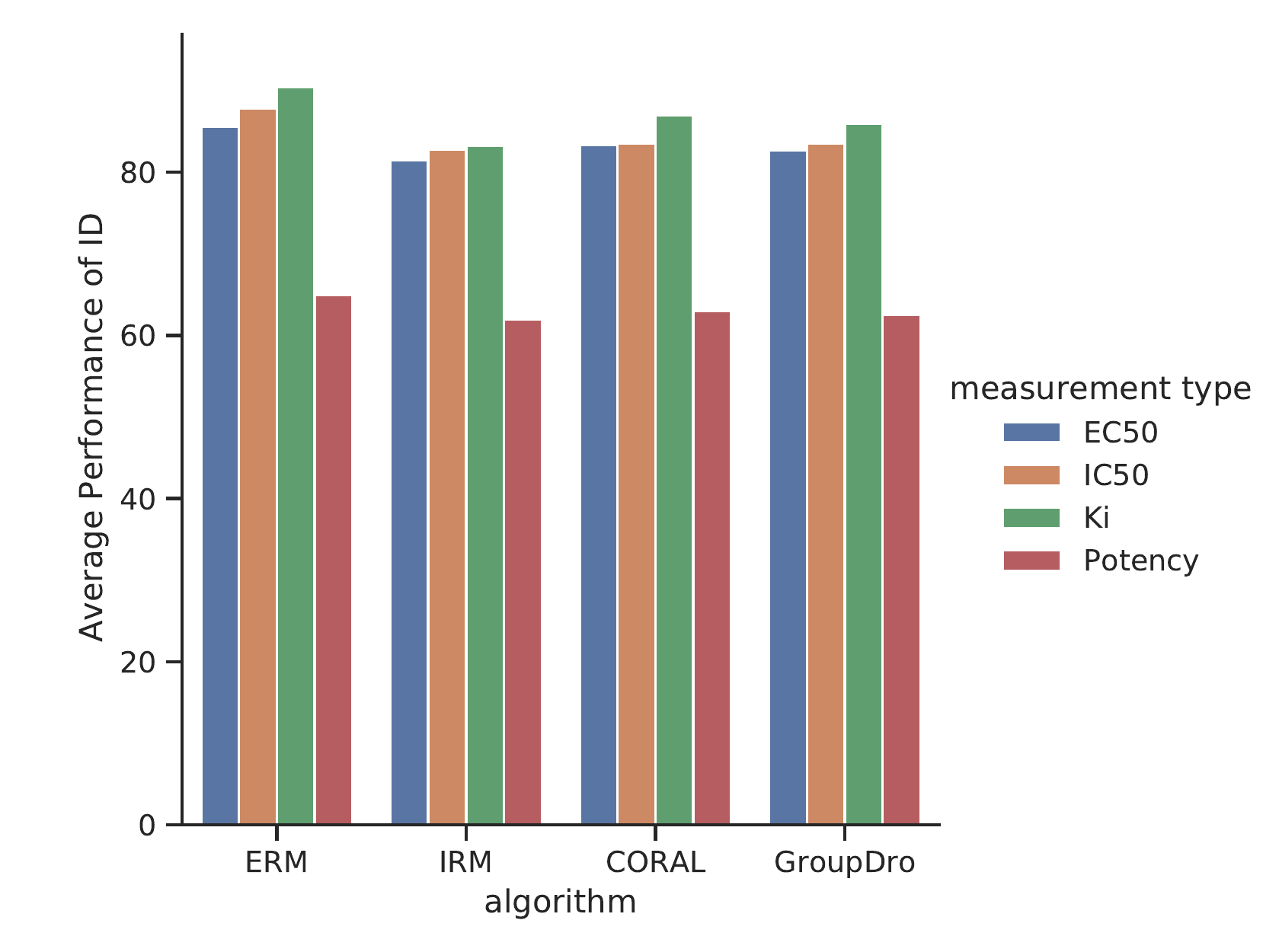}{0.35}{(b)}
\scalebarimg{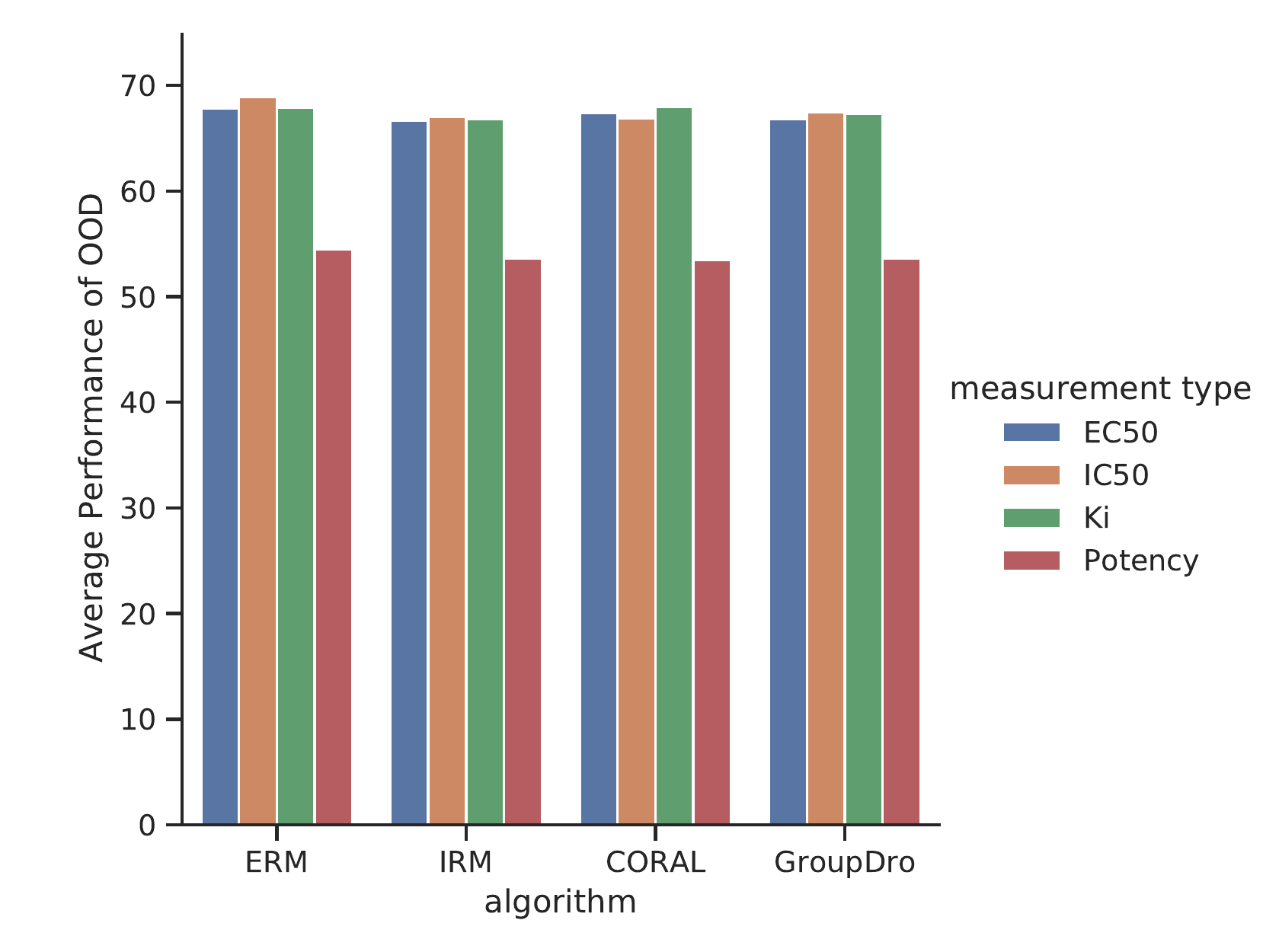}{0.35}{}
\caption{Average performance of ID (left) and OOD (right) for different measurement types. We present the performance of varying measurement types with the change of noise levels (a) and OOD algorithms (b).}
\label{fig:performance_of_measurement_types}
\end{figure}

\subsubsection{Study for Different Measurement Types}

The automated  dataset curator  supports variant measurement types, e.g., EC50, IC50. 
As different measurement types will generate datasets with different distributions and noise, we analyze the performance of different measurement types by varying the noise levels and algorithms.
The results are shown in ~\cref{fig:performance_of_measurement_types}.
One can see that:
1) As a result of different measurement types, ID and OOD performance can differ, for example, the Potency measurement is generally low in both ID and OOD, possibly due to the high noise level in the Potency measurement.
2) Our benchmark algorithms are robust to measurement types since they have acceptable accuracy for almost all types of measurement.

\section{Discussions and Future Work}
\label{sec_disc}

In this work we have presented an automated dataset curator and benchmark based on the large-scale bioassay deposition website \chembl, in order to facilitate OOD research for AI-aided drug discovery.  It is very worthwhile to explore more in the following respects.  

As observed in current benchmark results, existing general OOD methods do not significantly outperform the baseline ERM method. Most of these OOD methods are designed and validated with visual and/or textual data, which may fail in capturing critical information for the affinity prediction problem. This implies that to further improve the performance under various out-of-distribution scenarios, it is essential to develop more advanced OOD methods, particularly with drug-related domain knowledge integrated.

Another key characteristic of \drugood{} database is that the majority of data falls into the highest noise level (``general''). Simply discarding such noisy labels and only referring to high-quality ones may severely limit the model performance due to insufficient training data. It could be worth investigating that whether large-scale unsupervised pre-training can be utilized to construct better representations for molecules and target proteins, which are critical to accurate affinity predictions.

Additionally, learning with noisy labels has been extensively studied in the general context, but it may be crucial to take the generation process of noisy affinity annotations into consideration. This includes different experimental precision, measurement types, activity relation annotation types, etc. It is possible that the data quality can be further improved with carefully-designed denoising techniques, so that more accurate affinity prediction models can be trained.

\clearpage
{
\bibliography{bibs/bib_others,bibs/bib_noise,bibs/bib_ood,bibs/bib_general_drugai,bibs/bib_dataset_benchmark}
}

\clearpage 
\appendix
\appendixtitle{Appendix of \drugood}

\section{Statistics of the Realized Datasets}

\setlength{\tabcolsep}{1.3pt}
\renewcommand{\arraystretch}{1}
\begin{table}[!htp]
\centering
\caption{List of 36 lbap datasets in \drugood. Pos and Neg denote the numbers of positive and negative data points, respectively. $D^{\#}$ represents the number of domains, and $C^{\#}$ represent the number of data points.}
\scriptsize
\begin{tabular}{lccccccccccccc}\hline
\multirow{2}{*}{Data subset} &\multirow{2}{*}{$Pos^{\#}$} &\multirow{2}{*}{$Neg^{\#}$} &\multicolumn{2}{c}{Train} &\multicolumn{2}{c}{ID Val} &\multicolumn{2}{c}{ID Test} &\multicolumn{2}{c}{OOD Val} &\multicolumn{2}{c}{OOD Test} \\\cmidrule{4-13}
& & &$D^{\#}$ &$C^{\#}$ &$D^{\#}$ &$C^{\#}$ &$D^{\#}$ &$C^{\#}$ &$D^{\#}$ &$C^{\#}$ &$D^{\#}$ &$C^{\#}$ \\ \hline \hline
\drugood-lbap-core-ic50-assay &83802 &11434 &311 &34179 &311 &11314 &311 &11683 &314 &19028 &699 &19032 \\
\drugood-lbap-core-ic50-scaffold &83802 &11434 &6881 &21519 &1912 &4920 &24112 &30708 &6345 &19041 &4350 &19048 \\
\drugood-lbap-core-ic50-size &83802 &11434 &190 &36597 &140 &12153 &229 &12411 &4 &17660 &18 &16415 \\
\hline
\drugood-lbap-core-ec50-assay &10199 &2462 &47 &4540 &47 &1502 &47 &1557 &46 &2572 &101 &2490 \\
\drugood-lbap-core-ec50-scaffold &10200 &2462 &850 &2570 &224 &580 &3668 &4447 &1193 &2532 &953 &2533 \\
\drugood-lbap-core-ec50-size &10200 &2462 &167 &4684 &103 &1513 &205 &1753 &4 &2313 &17 &2399 \\
\hline
\drugood-lbap-core-ki-assay &22851 &488 &112 &8393 &112 &2769 &112 &2900 &83 &4631 &169 &4646 \\
\drugood-lbap-core-ki-scaffold &22851 &488 &1820 &5323 &521 &1192 &5826 &7490 &1366 &4665 &779 &4669 \\
\drugood-lbap-core-ki-size &22851 &488 &89 &8481 &62 &2799 &118 &2941 &4 &4644 &24 &4474 \\
\hline
\drugood-lbap-core-potency-assay &12265 &10554 &9 &8549 &9 &2848 &9 &2856 &7 &4008 &65 &4558 \\
\drugood-lbap-core-potency-scaffold &12032 &10787 &2003 &3361 &284 &433 &8910 &9898 &2125 &4563 &1509 &4564 \\
\drugood-lbap-core-potency-size &11994 &10825 &48 &8629 &38 &2859 &60 &2938 &3 &4182 &18 &4211 \\
\hline
\drugood-lbap-refined-ic50-assay &240996 &25526 &1446 &95373 &1446 &31401 &1446 &33164 &1361 &53293 &2805 &53291 \\
\drugood-lbap-refined-ic50-scaffold &240996 &25526 &19285 &60908 &5804 &14068 &66187 &84942 &15368 &53300 &9333 &53304 \\
\drugood-lbap-refined-ic50-size &240996 &25526 &225 &97779 &176 &32531 &264 &32847 &4 &51443 &20 &51922 \\
\hline
\drugood-lbap-refined-ec50-assay &32454 &5457 &141 &13600 &141 &4504 &141 &4655 &186 &7586 &372 &7566 \\
\drugood-lbap-refined-ec50-scaffold &32454 &5457 &2834 &7775 &745 &1647 &10945 &13324 &3082 &7587 &2187 &7578 \\
\drugood-lbap-refined-ec50-size &32454 &5457 &220 &13762 &140 &4524 &262 &4843 &5 &8558 &17 &6224 \\
\hline
\drugood-lbap-refined-ki-assay &72484 &2472 &578 &26752 &578 &8765 &578 &9462 &414 &14988 &905 &14989 \\
\drugood-lbap-refined-ki-scaffold &72484 &2472 &5643 &17172 &1692 &3899 &18612 &23914 &4436 &15013 &2386 &14958 \\
\drugood-lbap-refined-ki-size &72484 &2472 &148 &28230 &102 &9366 &189 &9591 &4 &13060 &23 &14709 \\
\hline
\drugood-lbap-refined-potency-assay &25382 &22611 &10 &17894 &10 &5962 &10 &5974 &7 &8899 &105 &9264 \\
\drugood-lbap-refined-potency-scaffold &25120 &22873 &4928 &8380 &850 &1145 &16819 &19270 &4344 &9598 &2927 &9600 \\
\drugood-lbap-refined-potency-size &25239 &22754 &51 &17951 &43 &5969 &65 &6042 &4 &10731 &18 &7300 \\
\hline
\drugood-lbap-general-ic50-assay &476865 &91691 &6917 &201951 &6917 &65424 &6917 &73777 &6207 &113704 &16814 &113700 \\
\drugood-lbap-general-ic50-scaffold &476865 &91691 &43552 &129740 &13516 &29174 &142173 &182220 &29513 &113723 &15189 &113699 \\
\drugood-lbap-general-ic50-size &476865 &91691 &290 &217294 &243 &72349 &311 &72742 &4 &102544 &22 &103627 \\
\hline
\drugood-lbap-general-ec50-assay &92445 &18000 &1079 &39333 &1079 &12849 &1079 &14086 &1137 &22095 &2883 &22082 \\
\drugood-lbap-general-ec50-scaffold &92445 &18000 &8677 &23481 &2450 &4977 &30611 &37811 &7659 &22095 &4844 &22081 \\
\drugood-lbap-general-ec50-size &92445 &18000 &294 &42697 &238 &14151 &312 &14531 &4 &19301 &20 &19765 \\
\hline
\drugood-lbap-general-ki-assay &146212 &6533 &2251 &54062 &2251 &17399 &2251 &20187 &1753 &30555 &4281 &30542 \\
\drugood-lbap-general-ki-scaffold &146212 &6533 &12334 &34395 &3625 &7472 &39232 &49784 &8780 &30559 &4280 &30535 \\
\drugood-lbap-general-ki-size &146212 &6533 &190 &55386 &140 &18407 &235 &18687 &5 &32508 &22 &27757 \\
\hline
\drugood-lbap-general-potency-assay &26784 &22389 &10 &17746 &10 &5912 &10 &5926 &9 &10146 &184 &9443 \\
\drugood-lbap-general-potency-scaffold &25617 &23556 &5052 &8647 &887 &1196 &17122 &19660 &4565 &9834 &2674 &9836 \\
\drugood-lbap-general-potency-size &25674 &23499 &51 &18224 &44 &6060 &66 &6138 &4 &10916 &18 &7835 \\
\hline
\end{tabular}
\label{tab:data_statistics_lbap_task:b}
\end{table}

\setlength{\tabcolsep}{0.5pt}
\renewcommand{\arraystretch}{0.8}
\begin{table}[!htp]
\centering
\caption{List of the 60 SBAP datasets in \drugood. Pos and Neg denote the numbers of positive and negative data points, respectively. $D^{\#}$ represents the number of domains and $C^{\#}$ represent the number of data points.}
\scriptsize
\begin{tabular}{lccccccccccccc}\hline
\multirow{2}{*}{Data subset} &\multirow{2}{*}{$Pos^{\#}$} &\multirow{2}{*}{$Neg^{\#}$} &\multicolumn{2}{c}{Train} &\multicolumn{2}{c}{ID Val} &\multicolumn{2}{c}{ID Test} &\multicolumn{2}{c}{OOD Val} &\multicolumn{2}{c}{OOD Test} \\\cmidrule{4-13}
& & &$D^{\#}$ &$C^{\#}$ &$D^{\#}$ &$C^{\#}$ &$D^{\#}$ &$C^{\#}$ &$D^{\#}$ &$C^{\#}$ &$D^{\#}$ &$C^{\#}$ \\ \hline \hline
\drugood-sbap-core-ic50-assay &104907 &15591 &351 &43250 &351 &14327 &351 &14737 &381 &24135 &723 &24049 \\
\drugood-sbap-core-ic50-protein &104907 &15591 &127 &43439 &127 &14449 &127 &14593 &134 &24006 &416 &24011 \\
\drugood-sbap-core-ic50-protein-family &104907 &15591 &1 &46890 &1 &15630 &1 &15631 &2 &21112 &10 &21235 \\
\drugood-sbap-core-ic50-scaffold &104907 &15591 &9911 &31201 &2560 &7149 &24334 &33948 &6151 &24138 &4322 &24062 \\
\drugood-sbap-core-ic50-size &104907 &15591 &192 &46390 &154 &15413 &229 &15685 &4 &21422 &18 &21588 \\
\hline
\drugood-sbap-core-ec50-assay &12053 &3287 &55 &5541 &55 &1831 &55 &1898 &53 &3023 &111 &3047 \\
\drugood-sbap-core-ec50-protein &12053 &3287 &29 &5524 &29 &1834 &29 &1867 &28 &3109 &66 &3006 \\
\drugood-sbap-core-ec50-protein-family &12053 &3287 &2 &5684 &2 &1894 &2 &1897 &2 &3124 &5 &2741 \\
\drugood-sbap-core-ec50-scaffold &12053 &3287 &1381 &3676 &292 &761 &3680 &4768 &1154 &3068 &980 &3067 \\
\drugood-sbap-core-ec50-size &12053 &3287 &170 &5772 &109 &1874 &205 &2130 &4 &2740 &17 &2824 \\
\hline
\drugood-sbap-core-ki-assay &31292 &852 &164 &11540 &164 &3801 &164 &3994 &116 &6427 &180 &6382 \\
\drugood-sbap-core-ki-protein &31292 &852 &37 &11604 &37 &3858 &37 &3904 &48 &6409 &125 &6369 \\
\drugood-sbap-core-ki-protein-family &31292 &852 &2 &16504 &2 &5501 &2 &5503 &  &  &9 &4636 \\
\drugood-sbap-core-ki-scaffold &31292 &852 &2712 &8640 &786 &1983 &5799 &8670 &1386 &6456 &786 &6395 \\
\drugood-sbap-core-ki-size &31292 &852 &99 &11934 &67 &3946 &118 &4094 &4 &5951 &24 &6219 \\
\hline
\drugood-sbap-core-potency-assay &18665 &15997 &10 &12892 &10 &4295 &10 &4305 &9 &6296 &63 &6874 \\
\drugood-sbap-core-potency-protein &19369 &15293 &9 &12699 &9 &4230 &9 &4241 &7 &6568 &30 &6924 \\
\drugood-sbap-core-potency-protein-family &18846 &15816 &1 &15701 &1 &5233 &1 &5235 &1 &4379 &6 &4114 \\
\drugood-sbap-core-potency-scaffold &18665 &15997 &3460 &7891 &847 &1487 &9034 &11430 &2099 &6921 &1411 &6933 \\
\drugood-sbap-core-potency-size &18665 &15997 &52 &13093 &46 &4352 &60 &4414 &3 &6243 &18 &6560 \\
\hline
\drugood-sbap-refined-ic50-assay &302752 &37178 &1760 &121655 &1760 &40060 &1760 &42245 &1712 &68003 &2937 &67967 \\
\drugood-sbap-refined-ic50-protein &302752 &37178 &219 &122326 &219 &40715 &219 &40992 &239 &68065 &1025 &67832 \\
\drugood-sbap-refined-ic50-protein-family &302752 &37178 &1 &130685 &1 &43561 &1 &43563 &2 &70150 &12 &51971 \\
\drugood-sbap-refined-ic50-scaffold &302752 &37178 &28561 &89798 &7690 &20509 &65773 &93651 &14954 &67993 &8900 &67979 \\
\drugood-sbap-refined-ic50-size &302752 &37178 &225 &123320 &182 &41046 &264 &41357 &5 &77829 &19 &56378 \\
\hline
\drugood-sbap-refined-ec50-assay &35977 &6428 &177 &15205 &177 &5016 &177 &5239 &212 &8484 &384 &8461 \\
\drugood-sbap-refined-ec50-protein &35977 &6428 &47 &15372 &47 &5112 &47 &5168 &59 &8351 &195 &8402 \\
\drugood-sbap-refined-ec50-protein-family &35977 &6428 &2 &18027 &2 &6008 &2 &6011 &1 &4791 &8 &7568 \\
\drugood-sbap-refined-ec50-scaffold &35977 &6428 &3774 &9950 &890 &2051 &10441 &13442 &2854 &8482 &1957 &8480 \\
\drugood-sbap-refined-ec50-size &35977 &6428 &228 &15809 &150 &5202 &262 &5527 &4 &7763 &18 &8104 \\
\hline
\drugood-sbap-refined-ki-assay &136133 &6050 &688 &50921 &688 &16783 &688 &17628 &702 &28440 &1131 &28411 \\
\drugood-sbap-refined-ki-protein &136133 &6050 &125 &51209 &125 &17033 &125 &17193 &121 &28364 &427 &28384 \\
\drugood-sbap-refined-ki-protein-family &136133 &6050 &2 &76074 &2 &25358 &2 &25359 &  &  &11 &15392 \\
\drugood-sbap-refined-ki-scaffold &136133 &6050 &9870 &41972 &2878 &10685 &19006 &32687 &4194 &28531 &2009 &28308 \\
\drugood-sbap-refined-ki-size &136128 &6047 &164 &54637 &117 &18164 &190 &18401 &4 &24120 &21 &26853 \\
\hline
\drugood-sbap-refined-potency-assay &39841 &32727 &13 &26983 &13 &8991 &13 &9010 &7 &13315 &99 &14269 \\
\drugood-sbap-refined-potency-protein &39841 &32727 &12 &26411 &12 &8800 &12 &8818 &7 &15531 &41 &13008 \\
\drugood-sbap-refined-potency-protein-family &41827 &30741 &1 &31027 &1 &10342 &1 &10344 &1 &11239 &6 &9616 \\
\drugood-sbap-refined-potency-scaffold &39841 &32727 &7645 &17457 &1872 &3292 &17214 &22791 &4216 &14549 &2545 &14479 \\
\drugood-sbap-refined-potency-size &39840 &32726 &53 &26233 &49 &8729 &65 &8810 &4 &16635 &18 &12159 \\
\hline
\drugood-sbap-general-ic50-assay &514296 &69105 &6171 &207522 &6171 &67554 &6171 &74982 &5678 &116671 &12019 &116672 \\
\drugood-sbap-general-ic50-protein &514296 &69105 &252 &209978 &252 &69923 &252 &70236 &307 &116773 &2112 &116491 \\
\drugood-sbap-general-ic50-protein-family &514296 &69105 &1 &224277 &1 &74759 &1 &74759 &2 &122574 &12 &87032 \\
\drugood-sbap-general-ic50-scaffold &514296 &69105 &50852 &156612 &14093 &35397 &109140 &158031 &24515 &116698 &12113 &116663 \\
\drugood-sbap-general-ic50-size &514296 &69105 &281 &226190 &233 &75318 &301 &75688 &4 &102468 &22 &103737 \\
\hline
\drugood-sbap-general-ec50-assay &66827 &8867 &728 &26966 &728 &8805 &728 &9667 &769 &15127 &1605 &15129 \\
\drugood-sbap-general-ec50-protein &66827 &8867 &68 &27256 &68 &9068 &68 &9158 &95 &15134 &468 &15078 \\
\drugood-sbap-general-ec50-protein-family &66827 &8867 &2 &31076 &2 &10358 &2 &10361 &1 &12651 &10 &11248 \\
\drugood-sbap-general-ec50-scaffold &66827 &8867 &6777 &18581 &1779 &3923 &17150 &22913 &4503 &15168 &2962 &15109 \\
\drugood-sbap-general-ec50-size &66827 &8867 &286 &27814 &228 &9195 &301 &9538 &5 &16147 &20 &13000 \\
\hline
\drugood-sbap-general-ki-assay &264464 &14083 &3368 &98880 &3368 &32029 &3368 &36227 &2995 &55701 &5795 &55710 \\
\drugood-sbap-general-ki-protein &264464 &14083 &124 &100249 &124 &33386 &124 &33528 &164 &55743 &948 &55641 \\
\drugood-sbap-general-ki-protein-family &264464 &14083 &2 &145756 &2 &48585 &2 &48588 &  & &12 &35618 \\
\drugood-sbap-general-ki-scaffold &264464 &14083 &21901 &81854 &6745 &20026 &38438 &65253 &8186 &55786 &3797 &55628 \\
\drugood-sbap-general-ki-size &264464 &14083 &206 &104541 &159 &34782 &233 &35078 &4 &48850 &22 &55296 \\
\hline
\drugood-sbap-general-potency-assay &40174 &33094 &13 &26972 &13 &8987 &13 &9005 &8 &14494 &138 &13810 \\
\drugood-sbap-general-potency-protein &40174 &33094 &12 &26423 &12 &8804 &12 &8820 &7 &15582 &52 &13639 \\
\drugood-sbap-general-potency-protein-family &42339 &30929 &1 &31218 &1 &10406 &1 &10407 &1 &11240 &7 &9997 \\
\drugood-sbap-general-potency-scaffold &40174 &33094 &7724 &17648 &1894 &3333 &17338 &22980 &4236 &14728 &2551 &14579 \\
\drugood-sbap-general-potency-size &40174 &33094 &53 &26482 &49 &8813 &65 &8889 &4 &16820 &18 &12264 \\
\hline
\end{tabular}
\label{tab:data_statistics_sbap_task:b}
\end{table}

\setlength{\tabcolsep}{20pt}
\renewcommand{\arraystretch}{1}
\begin{table}[!t]
\scriptsize
\centering
\caption{The in-distribution (ID) vs. out of distribution (OOD) of \drugood lbap datasets trained with empirical risk minimization. The ID test datasets are drawn from the training data with the same distribution, and the OOD test data are distinct from the training data.}
\begin{tabular}{lccc} \hline 
Dataset & In-dist & Out-of-Dist & Gap \\ \hline\hline
\drugood-lbap-core-ic50-assay&89.62 (2.04) &71.98 (0.29) &17.64\\
\drugood-lbap-core-ic50-scaffold&87.15 (0.48) &69.54 (0.52) &17.60\\
\drugood-lbap-core-ic50-size&92.35 (0.15) &67.48 (0.47) &24.87\\
\drugood-lbap-core-ec50-assay&85.23 (0.41) &70.30 (1.45) &14.93\\
\drugood-lbap-core-ec50-scaffold&84.86 (0.27) &68.07 (0.68) &16.80\\
\drugood-lbap-core-ec50-size&91.47 (0.20) &64.73 (0.55) &26.74\\
\drugood-lbap-core-ki-assay&89.10 (3.28) &76.63 (1.64) &12.46\\
\drugood-lbap-core-ki-scaffold&83.53 (2.15) &74.11 (2.37) &9.42\\
\drugood-lbap-core-ki-size&94.73 (0.74) &71.56 (2.07) &23.17\\
\drugood-lbap-core-potency-assay&66.23 (0.50) &53.26 (0.95) &12.97\\
\drugood-lbap-core-potency-scaffold&61.83 (0.31) &55.30 (0.72) &6.53\\
\drugood-lbap-core-potency-size&67.30 (0.36) &56.84 (1.35) &10.46\\
\drugood-lbap-refined-ic50-assay&89.25 (0.64) &72.70 (0.00) &16.55\\
\drugood-lbap-refined-ic50-scaffold&86.23 (0.08) &70.45 (0.54) &15.78\\
\drugood-lbap-refined-ic50-size&91.31 (0.07) &68.74 (0.37) &22.58\\
\drugood-lbap-refined-ec50-assay&76.74 (0.50) &71.05 (1.91) &5.69\\
\drugood-lbap-refined-ec50-scaffold&81.10 (0.50) &66.22 (0.34) &14.89\\
\drugood-lbap-refined-ec50-size&87.67 (0.29) &62.39 (0.70) &25.28\\
\drugood-lbap-refined-ki-assay&90.52 (0.26) &72.48 (1.39) &18.04\\
\drugood-lbap-refined-ki-scaffold&82.89 (0.66) &70.11 (2.10) &12.79\\
\drugood-lbap-refined-ki-size&89.23 (1.23) &72.44 (0.85) &16.78\\
\drugood-lbap-refined-potency-assay&62.99 (0.23) &58.88 (0.51) &4.11\\
\drugood-lbap-refined-potency-scaffold&60.54 (0.14) &56.00 (1.29) &4.54\\
\drugood-lbap-refined-potency-size&64.94 (0.71) &58.13 (0.49) &6.81\\
\drugood-lbap-general-ic50-assay&85.19 (1.15) &69.88 (0.13) &15.32\\
\drugood-lbap-general-ic50-scaffold&85.15 (0.24) &67.55 (0.09) &17.60\\
\drugood-lbap-general-ic50-size&89.77 (0.08) &66.05 (0.32) &23.72\\
\drugood-lbap-general-ec50-assay&83.32 (0.60) &69.55 (0.44) &13.77\\
\drugood-lbap-general-ec50-scaffold&80.12 (0.44) &63.66 (0.37) &16.46\\
\drugood-lbap-general-ec50-size&86.22 (0.23) &62.67 (0.47) &23.55\\
\drugood-lbap-general-ki-assay&87.98 (1.34) &74.54 (0.46) &13.44\\
\drugood-lbap-general-ki-scaffold&81.05 (3.11) &70.35 (0.83) &10.69\\
\drugood-lbap-general-ki-size&89.51 (0.88) &71.29 (0.94) &18.22\\
\drugood-lbap-general-potency-assay&63.40 (0.39) &56.38 (0.72) &7.03\\
\drugood-lbap-general-potency-scaffold&61.56 (0.29) &56.57 (0.42) &4.99\\
\drugood-lbap-general-potency-size&65.15 (0.30) &57.32 (0.54) &7.83\\
\hline
\end{tabular}
\label{tab:exp:lbap:erm_drop:c}
\end{table}

\setlength{\tabcolsep}{16pt}
\renewcommand{\arraystretch}{1}
\begin{table}[!t]
\scriptsize
\centering
\caption{The in-distribution (ID) vs. out of distribution (OOD) of \drugood sbap datasets trained with empirical risk minimization. The ID test dataset are drawn from the training data with the same distribution, and the OOD test data are distinct from the training data.}
\begin{tabular}{lccc} \hline 
Dataset & In-dist & Out-of-Dist & Gap \\ \hline\hline
\drugood-sbap-core-ic50-assay&91.32 (0.40) &71.04 (0.51) &20.28\\
\drugood-sbap-core-ic50-protein&90.71 (0.29) &68.87 (0.53) &21.84\\
\drugood-sbap-core-ic50-protein-family&89.88 (1.44) &72.20 (0.14) &17.68\\
\drugood-sbap-core-ic50-scaffold&88.97 (0.13) &70.73 (0.34) &18.24\\
\drugood-sbap-core-ic50-size&92.83 (0.07) &67.17 (0.16) &25.66\\
\drugood-sbap-core-ec50-assay&87.57 (0.80) &75.67 (1.34) &11.89\\
\drugood-sbap-core-ec50-protein&87.21 (0.78) &69.62 (4.45) &17.59\\
\drugood-sbap-core-ec50-protein-family&92.67 (0.94) &56.01 (3.03) &36.66\\
\drugood-sbap-core-ec50-scaffold&87.60 (0.32) &69.26 (1.20) &18.34\\
\drugood-sbap-core-ec50-size&92.98 (0.30) &63.99 (1.36) &28.98\\
\drugood-sbap-core-ki-assay&88.34 (4.18) &63.17 (2.46) &25.16\\
\drugood-sbap-core-ki-protein&85.05 (4.48) &64.89 (1.95) &20.16\\
\drugood-sbap-core-ki-scaffold&85.96 (1.10) &70.03 (2.90) &15.93\\
\drugood-sbap-core-ki-size&92.76 (0.18) &69.68 (1.13) &23.09\\
\drugood-sbap-core-potency-assay&70.22 (0.24) &51.39 (0.45) &18.83\\
\drugood-sbap-core-potency-protein&70.72 (0.48) &54.62 (0.57) &16.10\\
\drugood-sbap-core-potency-protein-family&62.60 (3.47) &49.83 (1.99) &12.77\\
\drugood-sbap-core-potency-scaffold&63.36 (0.55) &56.87 (0.53) &6.49\\
\drugood-sbap-core-potency-size&66.94 (0.37) &58.36 (0.23) &8.59\\
\drugood-sbap-refined-ic50-assay&88.07 (1.48) &69.21 (0.29) &18.86\\
\drugood-sbap-refined-ic50-protein&86.87 (1.41) &69.51 (0.30) &17.36\\
\drugood-sbap-refined-ic50-protein-family&86.44 (3.07) &70.61 (0.42) &15.82\\
\drugood-sbap-refined-ic50-scaffold&87.16 (0.08) &70.10 (0.52) &17.06\\
\drugood-sbap-refined-ic50-size&90.22 (0.11) &65.23 (0.26) &24.99\\
\drugood-sbap-refined-ec50-assay&80.60 (1.69) &70.38 (1.38) &10.22\\
\drugood-sbap-refined-ec50-protein&81.62 (1.16) &73.45 (1.67) &8.17\\
\drugood-sbap-refined-ec50-protein-family&83.09 (0.73) &68.81 (1.62) &14.27\\
\drugood-sbap-refined-ec50-scaffold&82.44 (0.19) &68.45 (0.28) &13.99\\
\drugood-sbap-refined-ec50-size&87.20 (0.04) &64.56 (0.51) &22.64\\
\drugood-sbap-refined-ki-assay&96.35 (0.56) &68.00 (1.73) &28.35\\
\drugood-sbap-refined-ki-protein&94.22 (0.48) &77.77 (0.60) &16.46\\
\drugood-sbap-refined-ki-scaffold&88.27 (0.88) &59.81 (1.04) &28.46\\
\drugood-sbap-refined-ki-size&91.56 (0.41) &61.64 (2.41) &29.92\\
\drugood-sbap-refined-potency-assay&64.65 (0.11) &54.66 (0.79) &9.99\\
\drugood-sbap-refined-potency-protein&65.46 (0.42) &53.14 (0.35) &12.32\\
\drugood-sbap-refined-potency-protein-family&64.76 (0.88) &53.75 (0.82) &11.01\\
\drugood-sbap-refined-potency-scaffold&61.90 (0.28) &55.05 (0.40) &6.85\\
\drugood-sbap-refined-potency-size&64.64 (0.65) &55.79 (0.26) &8.85\\
\drugood-sbap-general-ic50-assay&87.85 (1.47) &68.61 (0.54) &19.24\\
\drugood-sbap-general-ic50-protein&85.34 (1.67) &68.48 (0.27) &16.86\\
\drugood-sbap-general-ic50-protein-family&79.18 (2.69) &68.60 (0.68) &10.58\\
\drugood-sbap-general-ic50-scaffold&84.33 (0.67) &67.49 (0.33) &16.84\\
\drugood-sbap-general-ic50-size&87.81 (0.23) &65.34 (0.38) &22.47\\
\drugood-sbap-general-ec50-assay&84.66 (1.32) &69.83 (0.54) &14.82\\
\drugood-sbap-general-ec50-protein&82.10 (0.51) &70.16 (0.29) &11.94\\
\drugood-sbap-general-ec50-protein-family&83.69 (1.74) &63.76 (1.05) &19.93\\
\drugood-sbap-general-ec50-scaffold&82.97 (0.51) &67.95 (0.27) &15.01\\
\drugood-sbap-general-ec50-size&87.98 (0.42) &65.64 (0.35) &22.33\\
\drugood-sbap-general-ki-assay&94.15 (0.31) &72.05 (0.49) &22.10\\
\drugood-sbap-general-ki-protein&90.72 (0.14) &74.41 (0.25) &16.31\\
\drugood-sbap-general-ki-scaffold&87.09 (0.34) &67.36 (1.52) &19.73\\
\drugood-sbap-general-ki-size&91.10 (0.26) &65.93 (1.13) &25.17\\
\drugood-sbap-general-potency-assay&65.68 (0.17) &53.31 (0.32) &12.36\\
\drugood-sbap-general-potency-protein&64.83 (0.77) &53.35 (0.33) &11.48\\
\drugood-sbap-general-potency-protein-family&61.28 (0.87) &55.46 (1.24) &5.82\\
\drugood-sbap-general-potency-scaffold&61.91 (0.84) &56.02 (0.27) &5.89\\
\drugood-sbap-general-potency-size&65.80 (0.13) &55.33 (0.38) &10.47\\
\hline
\label{Implement:backbone}
\end{tabular}
\vspace{-1pt}
\label{tab:exp:sbap:erm_drop}
\end{table}

\end{document}